%% file: arxiv-wainer-v2.tex
\documentclass[twoside,11pt,preprint]{article}

\usepackage{jmlr2e}

\usepackage[utf8]{inputenc}
\usepackage[T1]{fontenc}
\usepackage{amsmath}
\usepackage{subfig}
\usepackage{floatrow}

\newfloatcommand{capbtabbox}{table}[][\FBwidth]
\def\tightlist{}
\usepackage{booktabs}
\usepackage{longtable}
\usepackage{array}
\usepackage{multirow}
\usepackage{wrapfig}
\usepackage{float}
\usepackage{colortbl}
\usepackage{pdflscape}
\usepackage{tabu}
\usepackage{threeparttable}
\usepackage{threeparttablex}
\usepackage[normalem]{ulem}
\usepackage{makecell}
\usepackage{xcolor}

\jmlrheading{1}{2022}{1-48}{4/00}{10/00}{wainer22a}{Jacques Wainer}

\ShortHeadings{BBT: Comparison of multiple algorithms on multiple data sets}{J. Wainer}
\firstpageno{1}

\begin{document}

\title{A Bayesian Bradley-Terry model to compare multiple ML algorithms on multiple data sets}

\author{\name Jacques Wainer \email \href{mailto:wainer@ic.unicamp.br}{\nolinkurl{wainer@ic.unicamp.br}}\\
       \addr Artificial Intelligence Lab, RECOD.ai\newline Institute of Computing\newline University of Campinas\newline Campinas 13083-970, Brazil.
       }

\editor{}

\maketitle

\begin{abstract}
This paper presents a Bayesian model, called the Bayesian Bradley Terry (BBT) model, for comparing multiple algorithms on multiple datasets based on any metric. The model is an extension of the Bradley Terry model, which tracks the number of wins each algorithm has on different datasets. Unlike frequentist methods such as Demšar tests on mean rank or Benavoli et al.'s multiple pairwise Wilcoxon tests, the Bayesian approach provides a more nuanced understanding of the algorithms' performance and allows for the definition of the ``region of practical equivalence'' (ROPE) for two algorithms.
Additionally, the paper introduces the concept of ``local ROPE,'' which assesses the significance of the difference in mean measure between two algorithms using effect sizes, and can be applied in frequentist approaches as well.
Both an R package and a Python program implementing the BBT are available for use.
\end{abstract}


\hypertarget{introduction}{%
\section{Introduction}\label{introduction}}

In the field of Machine Learning, new models or algorithms are often compared to existing ones using a variety of data sets. These comparisons usually result in a table \ref{tab:tabfirst}, where each line indicates a data set, and each column the algorithms being compared.

\begin{table}[ht]
\centering
\begin{tabular}{r|cccc}
 & Alg A & Alg B & ... & Alg L \\ \hline
 DB 1 & $a_1$ & $b_1$ & ... & $l_1$ \\
 DB 2 & $a_2$ & $b_2$ & ... & $l_2$ \\
 DB 3 & $a_3$ & $-$   & ... & $l_3$ \\
 ... & ... & ... & ... & ... \\
 DB K & $a_k$ & $b_k$ & ... & $l_k$ \\
\end{tabular}
\centering
\caption{The table of measures of comparing algorithms A, B, ..., L on datasets DB1, DB 2, ..., DB K.}\label{tab:tabfirst}
\end{table}

In the example above \(a_2\) is the measure of algorithm A on the data set DB 2. Usually, this measure is the average of a set of measures of algorithm A on different cross validations of the data set DB 2.

The different results will be compared using a \textbf{comparison procedure}.
The goals of a comparison procedure are in order of importance:

\begin{enumerate}
\def\labelenumi{\arabic{enumi}.}
\item
  The procedure should tell which algorithm is better, which is second place, which one
  is third place, and so on when measured on a particular set of data
  sets. We will call this the \textbf{aggregated ranking} of the
  algorithms for that set of data sets. In particular, for the example above, the aggregated ranking would result in a total order such as \(C \succ\,B \succ\,E \hdots A\) stating that C is better than B which is better than E, and so on, and that A is the worse algorithm in the set. We will indicate that algorithm C is better than algorithm B as \(C \succ\,B\). Of course, the meaning of better depends on the metric being used; higher numbers are better for metrics such as accuracy, F1, AUC, and lower numbers are better for metrics such as error, execution time, energy consumption and so on.
\item
  The procedure should compute how confident or how hopeful one should be that the ordering
  will remain true when one tests the same algorithms on a new
  data set. That is, the procedure should indicate the confidence in each of the comparisons \(C \succ\,B\), \(C \succ\,E\), \(C \succ\,A\), \(B \succ\,E\), and so on.
\item
  The procedure should state how much one algorithm is
  better than another or at least when one algorithm is not much
  better than another one, that is, when both algorithms are equivalent. Ideally one would like a measure dist(C,B) which indicates how much better C is from B. Or at least one would like such a distance measure that would indicate when C is not really much better than B and that, for practical purposes, they are equivalent.
\item
  The procedure should not require that all algorithms must be evaluated in all data sets. In the example above, algorithm B did not run for data set B, as indicated by the ``-'' entry in the table.
\end{enumerate}

There are numerous methods for aggregating rankings in machine learning. The obvious method of computing the mean of the measures of each algorithm and ordering them based on that mean is not considered appropriate for machine learning comparisons. There are two reasons for that. The first is that there are some metrics used in Machine Learning, and especially in regression tasks which are \textbf{non-comparable}. Two of such metrics are RMSE (root mean square error) and MAE (mean absolute error). Let us assume an algorithm that has an RMSE of \$30'000,00 in predicting housing values of Boston suburbs (DB 1) and an RMSE of 3.3 on predicting the quality of red wine (DB 2). How can one add those two numbers to obtain an average? In fact, we cannot even compare those two numbers; which one is higher? But even for comparable metrics, such as accuracy, AUC, and so on for classification, or MAPE (mean absolute percentage error) for regression, which are dimensionless quantities, there is a subtle problem in averaging those measures. For example, an improvement in accuracy from 76\% to 78\% is less ``significant'' than an improvement from 96\% to 98\%, even though both represent a 2\% increase. Two algorithms that have a mean accuracy of 0.86, but one with 78\% and 96\% accuracy on two datasets, and the other with 76\% and 98\% accuracy, would have the same mean accuracy, but the second algorithm would be considered ``better'' as its 2\% increase on the second dataset is more ``significant.'' Thus, computing the aggregated ranking based on the mean of the measures, even for comparable metrics, is not considered a ``correct'' procedure in Machine Learning.

Another alternative \citep{benavoli2016should, STAPOR2021107219} is to compute the median measure for each algorithm, provided the metric used is comparable. The aggregated ranking is then determined by ranking the medians.
One common method \citep{demsar} is to compute the rank of each algorithm within each data set, assigning 1 to the best, 2 to the second, and taking the average rank in case of ties. The mean rank of each algorithm is then calculated, and the ranking of these mean ranks determines the final aggregated rank.

Additionally, there are numerous ranking aggregation methods discussed in other disciplines \citep{langville2012s}, such as social choice theory where rankings are referred to as preferences and a wide range of voting procedures are used to aggregate them \citep{sep-social-choice}. However, it is important to note that these aggregation procedures in other disciplines are not usually associated with a measure of confidence in determining which item is ``better'' than another in the aggregated rank.

The second goal of evaluating the confidence of the aggregated ranking is only necessary when a sample of the relevant population is used for comparison. In the case of determining the winner of an election or a sports championship, the entire population is considered, and there is no need for a statistical evaluation of the victory. In these cases, the purpose of the comparison procedure is simply to calculate the aggregated ranking.

However, when comparing machine learning algorithms, the aggregated ranking obtained from a particular set of data sets is not the ultimate goal. The objective is to make claims about the ranking of algorithms on future, yet-to-be-seen data sets.
To assess the confidence in the aggregated ranking, statistical tests are often used to determine the trustworthiness of paired comparisons between algorithms. Historically, the most common approach has been the \textbf{frequentist} approach, where a null hypothesis significant test is used to make a binary decision on the significance of the difference between algorithms.

In recent times, there has been a shift towards \textbf{Bayesian} approaches to statistical testing. Bayesian approaches do not provide a binary decision on the significance of paired comparisons, but instead provide a probability of one algorithm being better than the other.

Regarding the third goal, there is a clear need to state that two
algorithms are similar for practical purposes. However, frequentist
methods are not well-suited to make this claim\footnote{There are frequentist
  tests that determine when two alternatives do not have practical
  difference. These are known as equivalence tests \citep{eqvtest} but are
  not of common use in Machine Learning or other areas of
  Computing.}. Some researchers mistakenly assume that if there is no
statistical difference between two algorithms, they are similar or
equivalent. However, a non-significant result from a frequentist test
only indicates that the sample size was not large enough to detect a
significant difference, not that the algorithms are equivalent. With a
large enough sample size, all p-values will go to 0.0 \citep{cosma, kruschke2015bayesian} and all differences will become significant. In
contrast, Bayesian methods do allow for the claim of practical
equivalence, as we will see in a later section (Section
\ref{sec:bayestests}).

Finally, regarding the fourth goal, it is desirable to have the
comparison procedure handle missing measures gracefully, as algorithms
may not converge on some data sets, may require more memory than is
available, or may exceed the allotted computational time. This is
particularly relevant in the case of frequentist approaches, as there
is no universally agreed-upon way of handling missing measures.

This paper presents a new comparison procedure that is based on the number of times an algorithm outperforms another on various data sets. The proposed procedure is as follows:

\begin{itemize}
\item
  The statistical framework used is the Bradley-Terry model for ranks, which assumes that each algorithm has a latent ``merit number'' or ``ability'' that determines the probability of it outperforming another algorithm.
\item
  The aggregated ranking of algorithms is determined by the ordering of these merit numbers.
\item
  The proposed procedure utilizes a Bayesian implementation of the Bradley-Terry model, which allows for the computation of the probability of one algorithm being better than another, and serves as a measure of confidence in the ordering.
\item
  The Bayesian model also enables the definition of when two algorithms are considered equivalent for practical purposes through the concept of the region of practical equivalence (ROPE).
\item
  The ROPE is defined in the probability space, allowing for a generic notion of equivalence that can be understood and modified by researchers, regardless of their experience with the particular metric being used.
\item
  A concept of local ROPE is also introduced, which is a decision criterion for comparing two algorithms on a specific data set. The decision is based not only on the difference between the two mean measures but also takes into consideration the ``noise level'' or effect size of the differences.
\end{itemize}

This paper is laid out as follows: Section \ref{sec:tut} is a short
tutorial on frequentist tests for multiple comparisons in general, and
Bayesian tests. A reader that knows the basics of frequentist tests can
skip section \ref{sec:tut1}; a reader that understands p-value
adjustments in multiple comparisons can skip Section \ref{sec:tut2}.

Section \ref{sec:prev}
discusses the previous frequentist and Bayesian approaches to
comparing multiple algorithms on multiple data sets.
Section \ref{sec:btmod} discusses this proposal, the use of a
Bayesian Bradley Terry model. Section \ref{sec:exp1} shows some first
results of using the BBT model. Section \ref{sec:lrope} discusses our
proposal of considering more than the mean across different
cross-validations when determining if one algorithm is better than the
other. Section \ref{compare-tests} compare the results of the BBT
model with the previous approaches. Section \ref{predictive} discusses
the quality of the BBT model as a predictive estimation of the future
behavior of the algorithms on new data sets. Section \ref{sec:disc}
discusses some of the advantages and shortcomings of the BBT model,
and section \ref{sec:conc} summarizes the main conclusions.

\hypertarget{a-short-tutorial-on-frequentist-and-bayesian-multiple-comparisons}{%
\section{\texorpdfstring{A short tutorial on frequentist and Bayesian multiple comparisons \label{sec:tut}}{A short tutorial on frequentist and Bayesian multiple comparisons }}\label{a-short-tutorial-on-frequentist-and-bayesian-multiple-comparisons}}

Let us assume that different algorithms tested on multiple data sets
are measured using some metric for which calculating the average is an
acceptable procedure (which is not the usual case in Machine Learning
results, as discussed above), and let us also assume that many data sets
were used in the comparison. These two assumptions are necessary for
this tutorial so: a) we can use a \textbf{parametric} frequentist test to
discover which algorithms are significantly different from each other
and b) use a \textbf{Gaussian} based Bayesian modeling of the problem.

\hypertarget{general-aspects-of-the-frequentist-approach}{%
\subsection{\texorpdfstring{General aspects of the frequentist approach \label{sec:tut1}}{General aspects of the frequentist approach }}\label{general-aspects-of-the-frequentist-approach}}

The frequentist approaches, also known as \textbf{NHST} or null hypothesis
statistical testing, will select a few statements of the form ``the
mean of results for algorithm A is different than the mean of results
for algorithm B'', or more succinctly that ``algorithm A is different
than algorithm B'' and claim that \emph{they are true} (but not in these
terms), and for the other possible statements, it will claim that it
cannot decide whether the statement is true or not true. More
formally, NHST will indicate that a difference between two
algorithms or groups is true by the statement ``the difference between algorithm
A and B is \emph{statistically significant}''. The test will indicate that a
difference cannot be shown to be true or false by the statement ``the
difference between algorithms A and B is \emph{not} statistically
significant''.

Internally, most frequentist test models the problem by assuming that
there is no difference between the populations that \emph{generated} the
samples, and computes an approximation of the probability that the
unique source of data for all groups would generate (by chance)
samples that have mean values as different, or more different, than the ones
encountered in the real data. This approximation of the probability
is called \textbf{p-value} and the standard in Machine Learning and most other Science areas is
that if \(\mbox{p-value} \le 0.05\) then one claims that
the null hypothesis (that all data came from a single population) is
false, and therefore, the ``differences between the algorithms are real''.

\hypertarget{frequentist-test---anova-plus-post-hoc-tests}{%
\subsection{\texorpdfstring{Frequentist test - ANOVA plus post-hoc tests \label{sec:tut2}}{Frequentist test - ANOVA plus post-hoc tests }}\label{frequentist-test---anova-plus-post-hoc-tests}}

When comparing the mean of multiple groups, the standards frequentist
approach is to perform an \textbf{omnibus} test first, followed, if needed,
by some \textbf{post hoc} tests. The omnibus test assumes a null
hypothesis that all groups came from the same population or
distribution. If the p-value of this test is low enough, then the
conclusion is that not all groups came from the same population, but
this does not tell us which groups are different from each other. This
will be the role of the post-hoc tests. In parametric multiple
comparison procedures, the omnibus test is called ANOVA (Analysis of
Variance) or repeated measures ANOVA, depending on whether the data is
not-paired or paired respectively.

\textbf{Paired} data refer to situations where there is a correspondence
among each data item in each of the groups. For example, each
algorithm was tested on the same set of data sets, and thus one can
make the correspondence of that result in each algorithm to one
particular data set. Usually, machine learning evaluations deals with
paired data, but that is not necessary. The alternative to paired data
is called \emph{independent} samples, or \emph{non-paired} data\footnote{In
  some statistical texts, the
  paired situation is also called \emph{within-subjects}, while the non-paired
  is called \emph{between-subjects}.}. Statistical tests differ for paired and
non-paired data, but one can always use a non-paired test on paired
data, but usually, with some decrease in power - some differences that
would be significant with the paired test may not be so using the
non-paired test.

Once the comparison passes the omnibus test, one performs the post-hoc
test, which verifies which of the claims ``group A is different than
group B'' has sufficient evidence. There are different ways of
classifying post-hoc tests, but for this paper we will divide pos-hoc tests into two group: the ones that compute a critical difference, and the ones that perform p-value adjustments. A critical difference test, for example Tukey's range test, computes a single critical threshold and if the difference between two group means is
larger than this threshold, then one can claim that the
difference between the two groups is statistically significant,
otherwise, the difference is not statistically significant. The
other family of post-hoc tests compute a separate p-value for each pair of groups, and adjust each of
the p-values, using some (statistical) \textbf{multiple comparisons} or
\textbf{p-value adjustment} procedure.

Statistical multiple comparison procedures are a complex topic in
frequentist statistics \citep{feise2002multiple, rothman1990no, bender2001adjusting}. We refer the reader to
\citep{farcomeni2008review, garcia2008extension} for a deeper
understanding of the issue and the techniques.
But in general terms, let us assume that each pairwise test uses the 0.05
threshold for p-value (this is called the
comparison-wise or individual error rate). If one makes \(k\)
\emph{independent} tests, and summarizes them into a single statement
``\(T_1\) and \(T_2\) and \(T_3\) and \(\ldots T_k\)'' the probability that any
one of the sub-statement (\(T_i\)) is correct is \(1-0.05\). If we assume
that the tests are independent, the probability that all
sub-statements are correct is \((1-0.05)^k\), and therefore, the
probability that the statement as a whole is false is \(1- (1-0.05)^k\).
This is called the family-wise error rate (FWER). If there are 100
tests, the FWER is 0.994, instead of the 0.05 one would expect from
each of the individual tests! If one wants the family-wise error to
be 0.05 one must use a much smaller individual error rate. On first
approximation, if the individual error rate is \(\alpha\), and if one wants the FWER to be 0.05, then:

\begin{align*}
1- (1-\alpha)^k  = 0.05\\
1 - (1^k -k 1^{k-1} \alpha + O(\alpha^2)) = 0.05\\
k \alpha = 0.05 - O(\alpha^2))\\
\alpha = \frac{0.05}{k}
\end{align*}

The calculations above reflect what is known as the Bonferroni test or Bonferroni
correction. The term \(O(\alpha^2)\) indicates that there is a term
with \(\alpha^2\), another with \(\alpha^3\) and so
on. Since \(\alpha\) is small, this calculation assumes that the
\(O(\alpha^2)\) term is very small and does not change the results in
any important way. Therefore, to maintain an FWER of 0.05, when
\(k=100\) each individual test should be accepted if its p-value is less
than 0.0005, that is \(0.05/100\). Statistical multiple test procedures
are usually expressed in terms of adjusting or modifying the original
p-values from each individual test so that one will still use the usual
0.05 threshold. In the example above, the Bonferroni procedure would
multiply each of the original p-values by 100 (limiting the product to
1.00 since p-value is a probability), and if any of the adjusted p-value
is lower than 0.05 then one accepts that comparison as statistically
significant.

There are problems with the assumptions for the Bonferroni correction:

\begin{itemize}
\item
  Not all tests are independent especially if one is comparing \(n\)
  different algorithms (or any items). In this case the \(k\) are all
  pairwise comparisons, that is \(k = \frac{n(n-1)}{2}\). If one test
  determines that algorithm A is better than B and B is better than
  C, then the third test, which compares A and C is fully
  determined, and it is not independent of the others. If, on the
  other hand, a test determines that A is better than B and C, the
  third test which compares B and C is indeed independent of the
  other two.
\item
  Not all \(k\) tests will return that a particular
  comparison/difference is statistically significant. If in the
  last example, the third test determines that there is not enough
  evidence to claim that B is better than C, the final claim will be
  ``A is better than B and A is better than C''. In this case, only
  the results of two tests were incorporated into the final claim.
\end{itemize}

The effective \(k\) one should use in the Bonferroni correction is lower
than the total number of tests and therefore the threshold p-value is
lower ``than it should be.'' This increases the type 2 error of the
procedure - some statements will not achieve the low p-value required
by the Bonferroni correction but would have a low enough p-value if the
correct \(k\) was used.

Beyond the Bonferroni procedure, there are dozen of different p-value
adjustment procedures, among them, Holm, Hochberg, Hommel, Benjamini and
Hochberg, and Benjamini and Yekutieli \citep{padj}. There are no clear
guidelines to select one or other p-adjustment procedures, but all are
more powerful than Bonferroni.

Because of the dependence on the number of comparisons, frequentist
approaches make a distinction between two goals of a
multiple comparisons. The first, and probably more common case, which
is called \textbf{comparisons against a control}, is the situation that
a researcher proposed a new algorithm and the researcher wants to
compare that algorithm, known as \emph{the control}, against others that are
considered competitors. The goal of the comparisons is to show
that the control is better or worse than each of its competitors and
the internal order among the competitors is not needed, nor it is
reported. The second goal, called \textbf{all pairwise
comparisons}, reflects a case where the researcher wants to rank a
set of algorithms on multiple data sets and all pair comparisons
are of interest and are reported. When one is comparing a set of
algorithms against a control, much fewer comparisons must be made,
which will require a less severe adjustment of the p-values.

\hypertarget{bayesian-tests}{%
\subsection{\texorpdfstring{Bayesian tests \label{sec:bayestests}}{Bayesian tests }}\label{bayesian-tests}}

Bayesian tests (also known as \textbf{Bayesian estimation procedures})
compute the posterior joint distribution of some parameters of the model
that are important for the analysis. The simpler case to discuss is a
Bayesian version of the two samples (non-paired) t-test for the means.
A simple Bayesian test will model each set of data (X and Y) as
samples from two Gaussian distributions with mean \(\mu_X\) and \(\mu_Y\)
and with standard deviation \(\sigma_X\) and \(\sigma_Y\) respectively.
Additionally, the model assumes that \(\mu_X\) and \(\mu_Y\) are themselves sampled
from another Gaussian with mean \(\mu\) and standard deviation \(\sigma\),
while \(\sigma_X\) and \(\sigma_Y\) are sampled from a uniform
distribution between \(L\) and \(H\)\footnote{A more complex model for this
  problem was proposed by \citet{kruschke2013bayesian}.}. This is expressed by
the following notation:

\begin{align}
x_i & \sim \mbox{Normal}(\mu_X, \sigma_X) \nonumber\\
y_i & \sim \mbox{Normal} (\mu_Y, \sigma_Y)\nonumber\\
\mu_X & \sim \mbox{Normal} (\mu,\sigma) \nonumber\\
\mu_Y & \sim \mbox{Normal} (\mu,\sigma) \label{eq:mod2}\\
\sigma_X & \sim \mbox{Unif}(L,H)\nonumber\\
\sigma_Y & \sim \mbox{Unif}(L,H) \nonumber 
\end{align}

The variables of interest are called \textbf{parameters} and in this case,
are \(\mu_X\) and \(\mu_Y\) (and maybe \(\sigma_X\) and \(\sigma_Y\)).

The distributions \(\mbox{Normal} (\mu,\sigma)\) and \(\mbox{Unif}(L,H)\)
are called \textbf{hyper-priors}. The \textbf{hyper-parameters} \(\mu, \sigma, L\), and \(H\) are set externally so that the distributions for
the true data X and Y are likely. For example, if the measured
average of the data in X is 5 and the measured average of the data in
Y is 5.2, then the random variables \(\mu_X\) and \(\mu_Y\) should be
around 5, and thus the mean of the Gaussian from which both \(\mu_X\)
and \(\mu_Y\) are sampled should have mean (the \(\mu\) hyper-parameter)
of 5.

The choice of hyper-priors and the hyper-parameters \(\sigma\), \(L\), and \(H\), can significantly impact the results of a Bayesian estimation procedure. Narrow hyper-priors may result in limited range of values for the parameters, and the results may be driven more by these constraints than the actual data. On the other hand, overly wide hyper-priors may have little to no constraints on the parameters, leading to potential convergence issues in the MCMC algorithm (as discussed in section\textasciitilde{}\ref{sec:mcmc1}). The debate on the appropriate width of hyper-priors is known as the non-informative vs weakly informative vs strongly informative prior debate \citep{lemoine2019moving, gelman2017prior}.

In general terms, a Bayesian test will compute (or sample, as we will
see below) the posterior distribution of the parameters of interests,
in this case \(\mu_X\) and \(\mu_Y\) given the data. That is, we want to
compute:

\[ P(\mu_X,\mu_Y | M, X, Y) = K \; \:  P(X,Y| M, \mu_X, \mu_y) P(\mu_X, \mu_Y | M)\]

where \(X\) and \(Y\) are the data, and \(M\) is the model itself
(Equations \ref{eq:mod2}), and \(K\) is a constant that normalizes the
distribution \(P(\mu_X,\mu_Y| ...)\).

Once the posterior distribution \(P(\mu_X,\mu_Y | M, X, Y)\) is computed, one can use it to make probabilistic statements about the parameters of interest. For example, \(P(\mu_X > \mu_Y | M, X, Y)\) represents the probability that the mean of the distribution that generated the data \(X\) is greater than the mean of the distribution that generated the data \(Y\). If the probability is 0.8, then the researchers are 80\% confident that \(X\) comes from a population whose mean is greater than the population from which \(Y\) comes from. This type of probabilistic statement is different from the claims made using frequentist methods.

\hypertarget{rope}{%
\subsubsection{\texorpdfstring{ROPE \label{sec:rope}}{ROPE }}\label{rope}}

Another useful statement to derive from the posterior distribution is the probability that the difference between \(\mu_X\) and \(\mu_Y\) is of no practical consequence. If \(\delta\) is the value below which the difference is considered as insignificant, then the value of \(P(|\mu_X - \mu_Y| < \delta | M, X, Y)\) represents the probability that there is no important difference between \(\mu_X\) and \(\mu_Y\). In Bayesian analysis, the value \(\delta\) is referred to as the \textbf{region of practical equivalence (ROPE)}. Differences smaller than the ROPE are considered to be of no practical significance.

\hypertarget{mcmc-and-convergence}{%
\subsubsection{\texorpdfstring{MCMC and convergence \label{sec:mcmc1}}{MCMC and convergence }}\label{mcmc-and-convergence}}

Typically, Bayesian tests do not calculate the probability distribution
\(P(\mu_X,\mu_Y | M, X, Y)\) analytically, but instead use a method from the Markov Chain Monte Carlo (MCMC) family of algorithms to sample pairs \(\langle {\mu_X}_s,{\mu_Y}_s \rangle\) from that distribution, denoted as \(\langle {\mu_X}_s,{\mu_Y}_s \rangle \sim P(\mu_X,\mu_Y | M, X, Y)\). In this paper, \(s\) will be used as the index for samples generated from the MCMC algorithm. From sample set \(\langle {\mu_X}_s,{\mu_Y}_s \rangle\), determining the probability that \(\mu_X > \mu_Y\) simply involves counting the proportion of samples for which \({\mu_X}_s > {\mu_Y}_s\).

MCMC algorithms eventually converge to the target distribution, but in practice, the algorithm will run for a pre-determined number of steps, and \textbf{convergence diagnostics} are used to assess whether the samples generated are representative of the target distribution. A comprehensive discussion of convergence diagnostics can be found in \citet{mcmcdiag}.

It is important to run convergence diagnostics every time a Bayesian model is run, as they provide information on whether the samples generated by the algorithm are representative of the posterior distribution of the parameters, or if more steps of the MCMC algorithm need to be run.

\hypertarget{posterior-predictive-check}{%
\subsubsection{Posterior Predictive Check}\label{posterior-predictive-check}}

\textbf{Posterior predictive} diagnostics aim to assess the accuracy of the Bayesian model (i.e., the model given by Equations \ref{eq:mod2}) in representing the data.

Even if the MCMC algorithm converges and returns samples from the posterior distributions of the parameters, the model may still not correctly describe the data. This is where the posterior predictive check (PPC) comes in, to verify that the data generated by the model, using the posterior values of the parameters, is similar to the observed data.

In essence, when the parameters assume the ``correct values'' (as determined by the posterior distribution of the parameters), the model can be ``run forward'' to generate new values of \(x_{is}\).\footnote{This is a simplification to provide an intuitive understanding. In reality, the MCMC algorithm also samples from \$ P(X\_\{rep\}, Y\_\{rep\} \textbar{} M, X, Y)\$, where \(X_{rep}\) is the data generated from the model (\(M\)) given the real data (\(X\) and \(Y\)).} The distribution of \(x_{1s}\), for example, which is the generated data for the first observation, should be compared to the ``real data'' \(x_1\). If the model fits the data well, the real data should be very similar to the set of generated data.

Other methods for evaluating the fit of a model to the data, such as leave-one-out cross-validation approximations (e.g., WAIC (Watanabe-Akaike information criteria) \citep{watanabe2010asymptotic} and loo \citep{vehtari2017practical}), are also available. However, we will not go into detail about these approaches here but will use them later when discussing alternative modeling options.

\hypertarget{bayesian-tests-for-multiple-comparisons}{%
\subsubsection{Bayesian tests for multiple comparisons}\label{bayesian-tests-for-multiple-comparisons}}

The example discussed above involves a Bayesian test on two sets of data (X and Y). If one wants to compare multiple sets of data, can one repeat the Bayesian test for all pairs? For frequentist tests, repeating the test for all pairs of comparisons would require some p-value adjustment procedure. However, the issue of performing multiple Bayesian comparisons is still unclear. It has been suggested that if the Bayesian model is hierarchical or multilevel, there would be no problem with performing multiple comparisons \citep{gelman2012we}. A hierarchical model
contains a model step similar to the line \ref{eq:mod2} in the Bayesian model described
above where the two parameters are sampled from a single distribution.

To perform multiple comparisons, a hierarchical Bayesian model is needed where each of the parameters of interest \(\mu_X, \mu_Y, \mu_W, \ldots \mu_Z\) are sampled from a common distribution. This process is known as partial pooling or shrinkage and it helps to pool the different estimates of the parameters towards each other. A good example of such a hierarchical model is the Bayesian ANOVA \citep[ch.~19]{kruschke2014doing}.

\hypertarget{frequentist-versus-bayesian-approaches}{%
\subsection{Frequentist versus Bayesian approaches}\label{frequentist-versus-bayesian-approaches}}

There are many differences between frequentist and Bayesian
approaches. We will not discuss them in this paper, but we point the
reader to \citet{benavoli2017time} discussion on the limitations of the
frequentist approach.

\hypertarget{previous-frequentist-and-bayesian-approaches}{%
\section{\texorpdfstring{Previous Frequentist and Bayesian approaches \label{sec:prev}}{Previous Frequentist and Bayesian approaches }}\label{previous-frequentist-and-bayesian-approaches}}

In this section, we will discuss the previous approaches to comparing multiple algorithms on multiple data sets.

\hypertarget{demsars-procedure-mean-rank-plus-nemenyi-test-and-extensions}{%
\subsection{\texorpdfstring{Demsar's procedure (mean rank plus Nemenyi test) and extensions \label{sec:demsar}}{Demsar's procedure (mean rank plus Nemenyi test) and extensions }}\label{demsars-procedure-mean-rank-plus-nemenyi-test-and-extensions}}

The standard frequentist procedure for comparing multiple algorithms on multiple datasets was introduced by \citet{demsar}. The main steps of this comparison procedure are:

\begin{itemize}
\item
  Convert each dataset's results into ranks, with 1 being the best result and 2 being the second best, etc.
\item
  Treat ties as the average rank. For example, if two algorithms have the same measure on a dataset and they are the fourth best ranked algorithms, then they both receive a rank of 4.5, which is the average of 4 and 5.
\item
  Determine the final order by decreasing the mean rank across all datasets. In other words, an algorithm with a lower mean rank is considered to be better than an algorithm with a higher mean rank.
\item
  The confidence in this order is indicated by a pairwise binary statement on whether one algorithm is better than another, using the phrase ``the difference is statistically significant.''
\item
  The significance of the pair comparisons is calculated using the Friedman test, followed by the Nemenyi test. If the difference in mean rank between two algorithms is smaller than the critical difference, then the difference is \textbf{not} statistically significant.
\end{itemize}

\citet{demsar} also discusses a different comparison procedure when comparing
against a control.

\citet{garcia2008extension} proposed and tested different extensions to Demšar's procedure, including Shaffer's static and Bergmann-Hommel's procedures, which they claimed were stronger than the Nemenyi test. \citet{garcia2010advanced} proposed new omnibus tests, such as the Friedman aligned ranks and Quade tests, and tested other post-hoc procedures for p-value adjustments. The authors concluded that procedures such as Holm, Hochberg, Hommel, Holland and Rom produce equivalent results.

In summary, \citet{demsar} and its extensions are a family of non-parametric
and paired multiple comparison procedures that are based on the rank
of the algorithms within each dataset. The omnibus procedure can be
the Friedman test or other tests proposed by \citet{garcia2010advanced}, and
the post-hoc procedures can be critical difference on the ranks
\citep{demsar, garcia2008extension} or Wilcoxon pairwise procedures on the
rank data, followed by various p-value adjustment procedures
\citep{garcia2010advanced}.

\hypertarget{pairwise-wilcoxon-plus-p-value-adjustment-procedures}{%
\subsection{\texorpdfstring{Pairwise Wilcoxon plus p-value adjustment procedures \label{sec:pairwil}}{Pairwise Wilcoxon plus p-value adjustment procedures }}\label{pairwise-wilcoxon-plus-p-value-adjustment-procedures}}

\citet{benavoli2016should} highlights a problem with comparison procedures based on mean ranks, that the results of the comparison between two algorithms can be dependent on the other algorithms being compared. To address this issue, they suggest using a pairwise Wilcoxon signed rank test between the measures obtained by each algorithm for all data sets, followed by an appropriate multiple comparisons adjustment procedure. This approach is also suggested by \citet{STAPOR2021107219}.

This procedure computes the aggregated ranking based on the median measure of each algorithm across the data sets. Of course, such procedure
require comparable metrics.

Before we advance, let us discuss the issues
regarding missing data in the frequentist approaches discussed so
far - the cases where an algorithm did not run for some data set. The
problem is that there is no standard way of dealing with the missing
data. For mean rank approaches such as Demsar's procedure one could
not compute the rank for the algorithm that did not run for a
particular data set, and not use this missing rank when computing the
mean rank, but that would imply that not running on a data set would
attribute to that algorithm its mean rank. This is possible but it is
not standard. For example \citet{fernandez2014we} did something similar but
that procedure was criticised by \citet{wainberg2016random} as adding a
positive bias to the algorithm that did not run on the data set. One could do the same
for the pairwise Wilcoxon tests, but again that would mean that an
algorithm that did not run on a data set receives its median measure
for that data set. We do not have an opinion on whether this is the
correct thing to do, or if one should assign to the algorithm the
worst rank; the problem is that there is no agreed upon way of dealing
with missing data for frequentist tests.

\hypertarget{bayesian-pairwise-signed-rank-test}{%
\subsection{\texorpdfstring{Bayesian pairwise signed-rank test \label{sec:bsr}}{Bayesian pairwise signed-rank test }}\label{bayesian-pairwise-signed-rank-test}}

\citet{benavoli2017time} developed two forms of a Bayesian version of the
Wilcoxon signed rank test (BSR), and argued for their use within the
Machine Learning practice. The first, simpler form uses a single measure per algorithm and data set, usually the mean of various measures obtained through cross-validation. The second form, which is more complex and referred to as the Bayesian hierarchical correlated t-test, uses each measure from each cross-validation fold in the computation. However, the authors suggest that practitioners should use the simpler form.

The model's parameter is the mean of the pairwise difference between
of the accuracy of the algorithms on the different data sets, and the model
computes a probability distribution for this parameter. The authors state
that if this parameter is between -0.01 and 0.01 (or a ROPE of 1\%)
there is no practical difference between the two algorithms being
compared. The justification for the 1\% ROPE \emph{for accuracy} is not
presented in the paper, but that number is not too different than the
ROPE threshold proposed by \citet{wainer2016} and twice as large than the one proposed by
\citet{WAINER2021115222}, based on different sets of empirical evidence.

On the data level, as we discussed above, a 1\% of difference may or
may not be important, depending on the accuracy itself. A 1\% change for a
79\% accuracy is likely insignificant but a 1\% change for an accuracy
of 98\% is impressive. Given the range of accuracy values that appear
in practical cases of comparing two classifiers (some high, some low)
the authors are claiming that changes on the \emph{mean value} of less than
1\% are irrelevant from a practical point of view.

Even if one accepts the 0.01 ROPE for accuracy, there is no agreed
upon, or even proposed (as far as this author is aware) ROPEs for other classification metrics such as AUC, F1, MCC, and for comparable
regression metrics. Additionally, there is no ROPE for
incomparable metrics, nor will the Bayesian signed rank method be
applicable to incomparable metrics, given that the mean of the
pairwise differences is not conceptually well defined.

The Bayesian signed rank test was defined for comparing \emph{two}
algorithms on multiple data sets, and there may be issues when applying it to multiple comparisons. \citet{benavoli2017time} do not perform
any multiple comparison in the paper. They do perform many
signed-rank procedures with different algorithms but not with the goal
of ranking those algorithms. The paper
acknowledge that the Bayesian signed rank model lacks a
hierarchical component and therefore its use in multiple comparisons is problematic.\\
However, they argue that using ROPEs would mitigate the false alarm rate (the rate of false positive claims), but this argument is not widely accepted and would require the use of ROPEs for all comparisons. Since the authors only propose a ROPE for accuracy, it would be imprudent to use the Bayesian signed rank test for other metrics without further proposals for ROPEs for those metrics.

\hypertarget{bradley-terry-model}{%
\section{\texorpdfstring{Bradley-Terry model \label{sec:btmod}}{Bradley-Terry model }}\label{bradley-terry-model}}

The Bradley-Terry (BT) model \citep{bradley1952rank} is a method for ranking
``players'' in ``tournaments'' where the payers compete pairwise in
matches, such as soccer teams or chess payers. The model
assigns to each player \(X_i\) an \emph{intrinsic value} or \emph{ability}
\(w_i\). The intrinsic value relates to the probability that player \(i\)
will win player \(j\) in a match by:

\[ P(X_i \mbox{ wins } X_j) = P(X_i \succ\,X_j) = \frac{w_i}{w_i + w_j} \]

The final ranking of the players is defined by the rank of their
intrinsic values \(w_i\).

The intrinsic values are invariant to a multiplicative constant, that
is, if the set \(\{w_i\}\) correctly models all the probabilities \(P(a \succ\,b)\), so will \(\{\alpha w_i\}\). Therefore, to specify a single
set of intrinsic values, one also requires that \(\sum w_i = 1\).

An alternative to the \(w_i\) is to use their natural logarithms \(\beta_i = \log w_i\).
The useful formula regarding \(\beta\) is:

\begin{equation*}
\mbox{logit}(i \succ\,j) = \log \frac{P(i \succ\,j)}{1-P(i \succ\,j)} = \log \frac{P(i \succ\,j)}{P(j \succ\,i)} = \beta_i - \beta_j 
\end{equation*}

The \(\beta\) values are invariant to additive constant, since the \(w_i\)
were invariant to multiplicative constants. To specify a single set of solutions, a common practice is to require that \(\sum \beta_i = 0\).

The standard BT model does not deal with ties, meaning that \(1 - P(i \succ\,j) = P(j \succ\,i)\). However, there have been extensions to the model that incorporate ties \citep{rao1967ties, davidson1970extending, baker2021modifying}. The model proposed by \citet{davidson1970extending} will be discussed in Section \ref{test-ties}.

Inthe case of ties, a common approach is to change the data used for estimation so that a tie between players \(i\) and \(j\) is counted as both a victory for \(i\) and for \(j\) or sometimes as half a victory for each. This will be further explained in Section \ref{test-ties}.

\hypertarget{mle-and-bayesian-estimation-of-the-w-or-beta}{%
\subsubsection{\texorpdfstring{MLE and Bayesian estimation of the \(w\) or \(\beta\) \label{sec:bayesbt}}{MLE and Bayesian estimation of the w or \textbackslash beta }}\label{mle-and-bayesian-estimation-of-the-w-or-beta}}

Let us assume that players \(i\) and \(j\) play \(N_{ij}= N_{ji}\) matches against each other,
and \(W_{ij}\) is the number of matches that \(i\) wins and \(W_{ji}\) is
the number of matches \(j\) wins (and thus \(N_{ij} = W_{ij} + W_{ji}\)
given that there are no ties). If we assume that \(W_{ii} = 0\), and if
there are \(t\) players, then the likelihood function is:

\[L(w) = \prod_{i,j=1}^t (\frac{w_i}{w_i+w_j } )^{W_{ij} }\]

Given all the \(W_{ij}\), the maximum likelihood (MLE) estimation is the set \(w^* = \{w_1, w_2, \ldots w_n \}\) such that

\[w^* = \mbox{argmax}_w L(w)\]

There has been proposals for different algorithms to compute the MLE
(and MAP) solutions for the BT model
\citep{hunter2004mm, caron2012efficient}. Some research refer to the
case where the comparison graph is sparse \citep{li2021, butler2004},
which is not the case in our problem -- we assume that most of the
algorithms will be compared to almost all other algorithms on most
data sets.

The Bayesian model for BT is based on the beta coefficients:

\begin{align} 
 W_{ij} &\sim \mbox{Binomial}(N_{ij}, \frac{e^{\beta_i}}{e^{\beta_i} +  e^{\beta_j}}) \label{eq:mod1} \\
 \beta_i &\sim \mbox{Normal}(0,\sigma) \nonumber \\
 \sigma &\sim \mbox{LogNormal}(0,0.5) \nonumber
\end{align}

The binomial expression captures the fact that the number of times \(i\)
wins from \(j\) is a binomial distribution given the total number of
matches between \(i\) and \(j\) (\(N_{ij}\)) and the probability that \(i\)
will win each match (\(P(i \succ\,j) = \frac{w_i}{w_i+ w_j})\).

The \(\beta\) parameters can have positive and negative values and thus
it is reasonable to sample them from a normal distribution with mean
0, and variance \(\sigma\) (a hyper-parameter). This is the
hierarchical component of the Bayesian BT model (BBT): all \(\beta_i\)
are sampled from the same distribution and thus the model can be used
to compare multiple algorithms, since there will be partial pooling.
The hyper-prior for \(\sigma\) is a log-normal distribution, as proposed
by \citet{btstan}, but in section \ref{test-hyper} we explore other
hyper-priors and show that there is no difference on whether one uses
them.

\hypertarget{exploration-of-the-bbt}{%
\section{\texorpdfstring{Exploration of the BBT \label{sec:exp1}}{Exploration of the BBT }}\label{exploration-of-the-bbt}}

This section presents an analysis of the BBT model applied to a specific set of algorithms and data sets (detailed in Section \ref{sec:data}). The outputs of the model will be discussed in Section \ref{sec:out1}. Additionally, two forms of diagnostic checks are examined in Sections \ref{sec:conv1} and \ref{sec:ppc1}, as well as the concept of ROPE appropriate for the model in Section \ref{sec:rope2}. The section concludes with a discussion of the two interpretations of the parameters of the model in Section \ref{strong-weak}.

\hypertarget{data}{%
\subsection{\texorpdfstring{Data \label{sec:data}}{Data }}\label{data}}

We will explore the use of the BBT model on four use-cases regarding
the comparison of machine learning algorithms on multiple data sets.
The four use cases are called small-small, small-large, medium-medium,
and large-large.

The large-large ($\ell\ell$) use-case involves the evaluation of 16 out-of-the-box classifiers on 132 data sets. These classifiers were trained without any tuning of their hyper-parameters and the data sets are the first 132 smallest data sets from the PMLB dataset curated by \citet{Olson2017PMLB}. The accuracy metric was used for comparison with the BSR procedure.

The first set of 13 algorithms are implementations of classification algorithms available in the scikit-learn package \citep{scikit-learn} (version 1.1.1) with their respective default hyper-parameter values, which are not listed here.

\begin{itemize}
\tightlist
\item
  \emph{dt}: Decision tree
\item
  \emph{gbm}: Gradient boosting classifier,
\item
  \emph{knn}: K-nearest neighbors
\item
  \emph{lda}: Linear discriminant analysis,
\item
  \emph{lr}: Logistic regression
\item
  \emph{mlp}: Multi-layer perceptron
\item
  \emph{nb}: Naive Bayes Gaussian classifier
\item
  \emph{passive}: Passive-aggressive classifier,
\item
  \emph{qda}: Quadratic discriminant analysis,
\item
  \emph{rf}: Random forest
\item
  \emph{ridge}: Ridge regression
\item
  \emph{svm}: SVM with a RBF kernel
\item
  \emph{svml}: SVM with linear kernel
\end{itemize}

The next two algorithms are implemented in the XGBoost package \citep{Chen:2016:XST:2939672.2939785} (version 1.6.1):

\begin{itemize}
\tightlist
\item
  \emph{xgb}: Gradient boosting classifier as implemented by the XGBoost package
\item
  \emph{xrf}: Random forest as implemented by the XGBoost package
\end{itemize}

And the final algorithm is from the LightGBM package \citep{ke2017lightgbm} :

\begin{itemize}
\tightlist
\item
  \emph{lgbm}: Gradient boosting classifier as implemented by the LightGBM
\end{itemize}

The large-large results reflect the scenario where a large number of algorithms are compared on a large number of data sets. Most curated sets of data sets, such as PMLB \citep{Olson2017PMLB}, KEEL imbalanced data sets \citep{alcala2011keel}, and the OpenML-CC18 Curated Classification benchmark \citep{oml-benchmarking-suites}, include around 100 data sets. In this scenario, a researcher might typically compare around 20 algorithms, although there are some studies that test up to 50 \citep{wainer2021tune} or 100 \citep{fernandez2014we, wang2021comparing}. However, these studies do not perform statistical tests to confirm the statistical significance of their results within the frequentist framework or make any other probabilistic claims within a Bayesian framework. The $\ell\ell$-results are part of the R package developed for this research (Section \ref{sec:code}).

The other use cases are:

\begin{itemize}
\tightlist
\item
  the small-small (\emph{ss}) use-case, which represents the comparison of a small number of algorithms (5) on a small number of data sets (20).
\item
  the small-large (\emph{sl}) use-case which represents the comparison of 5 algorithms on 100 data sets.
\item
  the medium-medium (\emph{mm}) use case which represents the comparison of 10 algorithms on 50 data sets.
\end{itemize}

In this paper, the \emph{ss}, the \emph{sl}, and the \emph{mm} cases will be used in \emph{repeated experiments} to test some general claim regarding the BBT procedure, by sampling from the $\ell\ell$-results 10 random \emph{ss} and \emph{mm} results, and 5 random \emph{sl} results.

A fixed \emph{ss} result will be used to illustrate the BBT procedure
throughout this paper. This result is called the \emph{base} results. For the base result we
selected \emph{lgbm}, \emph{xgb}, \emph{svm}, \emph{lda}, and \emph{dt} as the classification
algorithms, and also selected 20 arbitrary data sets from the 132 in
the $\ell\ell$-results.

The table of values for the base results is displayed in Table \ref{tab:sstab1}.
The table represents the \emph{mean} accuracy on the \emph{same} 4-fold evaluation of the algorithms on each data-set.

\begin{table}

\caption{\label{tab:xsstab1}\label{tab:sstab1}The base results.}
\centering
\begin{tabular}[t]{lrrrrr}
\toprule
\textbf{db} & \textbf{dt} & \textbf{lda} & \textbf{lgbm} & \textbf{xgb} & \textbf{svm}\\
\midrule
\cellcolor{gray!6}{biomed} & \cellcolor{gray!6}{0.837} & \cellcolor{gray!6}{0.842} & \cellcolor{gray!6}{0.876} & \cellcolor{gray!6}{0.890} & \cellcolor{gray!6}{0.886}\\
breast & 0.931 & 0.951 & 0.964 & 0.961 & 0.957\\
\cellcolor{gray!6}{breast\_w} & \cellcolor{gray!6}{0.940} & \cellcolor{gray!6}{0.950} & \cellcolor{gray!6}{0.961} & \cellcolor{gray!6}{0.961} & \cellcolor{gray!6}{0.961}\\
buggyCrx & 0.790 & 0.861 & 0.867 & 0.867 & 0.861\\
\cellcolor{gray!6}{clean1} & \cellcolor{gray!6}{1.000} & \cellcolor{gray!6}{1.000} & \cellcolor{gray!6}{1.000} & \cellcolor{gray!6}{1.000} & \cellcolor{gray!6}{0.968}\\
\addlinespace
cmc & 0.455 & 0.513 & 0.525 & 0.524 & 0.544\\
\cellcolor{gray!6}{colic} & \cellcolor{gray!6}{0.761} & \cellcolor{gray!6}{0.837} & \cellcolor{gray!6}{0.815} & \cellcolor{gray!6}{0.815} & \cellcolor{gray!6}{0.641}\\
corral & 1.000 & 0.900 & 1.000 & 1.000 & 1.000\\
\cellcolor{gray!6}{credit\_g} & \cellcolor{gray!6}{0.668} & \cellcolor{gray!6}{0.718} & \cellcolor{gray!6}{0.766} & \cellcolor{gray!6}{0.769} & \cellcolor{gray!6}{0.724}\\
diabetes & 0.714 & 0.772 & 0.747 & 0.742 & 0.758\\
\addlinespace
\cellcolor{gray!6}{ionosphere} & \cellcolor{gray!6}{0.869} & \cellcolor{gray!6}{0.866} & \cellcolor{gray!6}{0.940} & \cellcolor{gray!6}{0.932} & \cellcolor{gray!6}{0.934}\\
irish & 1.000 & 0.740 & 1.000 & 1.000 & 0.988\\
\cellcolor{gray!6}{molecular\_b...y\_promoters} & \cellcolor{gray!6}{0.727} & \cellcolor{gray!6}{0.689} & \cellcolor{gray!6}{0.896} & \cellcolor{gray!6}{0.887} & \cellcolor{gray!6}{0.802}\\
monk3 & 0.975 & 0.792 & 0.980 & 0.986 & 0.964\\
\cellcolor{gray!6}{prnn\_crabs} & \cellcolor{gray!6}{0.880} & \cellcolor{gray!6}{1.000} & \cellcolor{gray!6}{0.950} & \cellcolor{gray!6}{0.935} & \cellcolor{gray!6}{0.960}\\
\addlinespace
prnn\_synth & 0.800 & 0.852 & 0.824 & 0.828 & 0.856\\
\cellcolor{gray!6}{saheart} & \cellcolor{gray!6}{0.626} & \cellcolor{gray!6}{0.723} & \cellcolor{gray!6}{0.660} & \cellcolor{gray!6}{0.671} & \cellcolor{gray!6}{0.712}\\
threeOf9 & 0.996 & 0.809 & 1.000 & 0.998 & 0.992\\
\cellcolor{gray!6}{tokyo1} & \cellcolor{gray!6}{0.902} & \cellcolor{gray!6}{0.920} & \cellcolor{gray!6}{0.928} & \cellcolor{gray!6}{0.926} & \cellcolor{gray!6}{0.931}\\
vote & 0.929 & 0.956 & 0.945 & 0.959 & 0.956\\
\bottomrule
\end{tabular}
\end{table}

\hypertarget{basic-outputs-of-the-model}{%
\subsection{\texorpdfstring{Basic outputs of the model \label{sec:out1}}{Basic outputs of the model }}\label{basic-outputs-of-the-model}}

A \textbf{win/loss table} is the representation of the number of wins and losses for each pair of algorithms.
This data is the input for the Bayesian model.

\begin{table}[ht]
\centering
\subfloat[The win/loss table for the base results. The win/loss table lists all win, ties, and losses for each pair of algorithms.]{\label{tab:wintab1}\scalebox{1.0}{\input{./twin1}}}\quad
\subfloat[The final win/loss table for the base results once the ties have been added as half-victories (rounded up in the final) to both algorithms.]{\label{tab:wintab2}\scalebox{1.0}{\input{./twin2}}}
\caption{The win/loss tables pre as post processing of the ties.}
\end{table}

Table \ref{tab:wintab1} displays some ties between algorithms, such as \emph{dt} and \emph{lda}
both having accuracy of 1.0 in the \texttt{clean1} data set. The BBT model does not handle ties. To address this, we implement the ``spread'' policy, which considers half (rounded up) of the ties as partial victories for both algorithms. The implementation of the spread policy is discussed in further detail in Section \ref{test-ties}. The resulting win/loss table is presented in Table \ref{tab:wintab2}.

The MCMC solution to the BBT model in Equations \ref{eq:mod2} for the data in
Table \ref{tab:wintab2} is a set of tuples for the parameters
\(\beta_{i}\), and for \(\sigma\). In our case, we are interested in
using, for example, each \(\beta_{as}\) and \(\beta_{bs}\) to compute
\(P_s(a \succ\,b)\). This computation is performed for all pairs of algorithms, and the results can be visualized or summarized in a plot or table.

The Bayesian approach to the BT model presents a challenge in determining the aggregated ranking of the algorithms. One commonly used solution is to calculate the ranking for each sample generated by the MCMC algorithm by ordering the algorithms based on their decreasing values of \(\beta_{is}\) \citep[\citet{issa2021bayesian}]{btstan}. However, this approach results in a distribution of rankings, making it unclear how to arrive at a single, final ordering. One can choose the most frequent ranking among the samples, or compute the rank of each algorithm in each ranking and order them based on the mean rank.

We believe that determining the order of the algorithms is a crucial part of the comparison process, so we propose a different solution. We order the algorithms based on their mean \(\beta\) across all samples, resulting in a single, aggregated ranking for the BBT comparison procedure.

\begin{figure}
\includegraphics[width=0.7\linewidth]{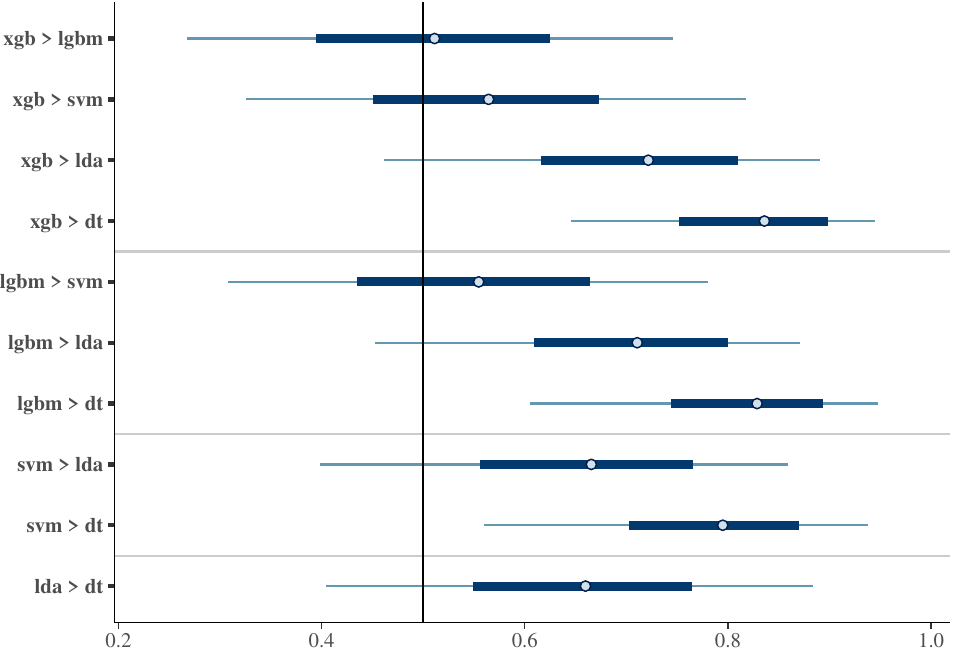} \caption{\label{fig:plot1}The graphical representation of the distribution of P(a > b). The pair of algorithms are ordered best > worse. The central dot represents the median of the distribution of P (a better than b), and the wider line represents the 89\% highest density interval (HDI) of the distribution. The thin line represents the full range of the distribution.}\label{fig:xplot1}
\end{figure}

Figure \ref{fig:plot1} presents the distributions of \(P_s(a \succ\,b)\). The algorithms are ordered from the best to the worst, where the best algorithm is compared with all others, the second best with the remaining worse, and so on. The central dot represents the median of the distribution of \(P_s(a \succ\,b)\), and the wider line represents the 89\% highest density interval (HDI) of the distribution. The thin line represents the full range of the distribution.

Some Bayesian estimation researchers use 89\% (instead of 95\%) to distinguish a credible interval, which is an interval that contains a specified amount of the mass of a distribution (in this case 89\%), from the ``95\% confidence interval'' concept in frequentist statistics, which has a slightly different meaning \citep{makowski2019bayestestr}. There are infinitely many intervals that contain 89\% of the mass of a distribution, and the HDI is the smallest of these intervals for unimodal distributions \citep{kruschke2014doing}.

Some of the information presented in Figure \ref{fig:plot1} can be condensed into Table \ref{tab:main1}, which includes the mean, as well as the low and high limits of the 89\% HDI.

\begin{table}

\caption{\label{tab:xresume1}\label{tab:main1}The table representation of the distributions of probabilities P(a > b). Pair is the pair of better/worse algorithms, mean is the mean probability, low and high are the lower and higher limits of the 89\% HDI of the distribution.}
\centering
\begin{tabular}[t]{lrrr}
\toprule
\textbf{pair} & \textbf{mean} & \textbf{low} & \textbf{high}\\
\midrule
\cellcolor{gray!6}{xgb > lgbm} & \cellcolor{gray!6}{0.51} & \cellcolor{gray!6}{0.40} & \cellcolor{gray!6}{0.63}\\
xgb > svm & 0.56 & 0.47 & 0.69\\
\cellcolor{gray!6}{xgb > lda} & \cellcolor{gray!6}{0.72} & \cellcolor{gray!6}{0.62} & \cellcolor{gray!6}{0.81}\\
xgb > dt & 0.83 & 0.76 & 0.90\\
\cellcolor{gray!6}{lgbm > svm} & \cellcolor{gray!6}{0.55} & \cellcolor{gray!6}{0.43} & \cellcolor{gray!6}{0.66}\\
\addlinespace
lgbm > lda & 0.71 & 0.62 & 0.81\\
\cellcolor{gray!6}{lgbm > dt} & \cellcolor{gray!6}{0.82} & \cellcolor{gray!6}{0.76} & \cellcolor{gray!6}{0.90}\\
svm > lda & 0.66 & 0.56 & 0.77\\
\cellcolor{gray!6}{svm > dt} & \cellcolor{gray!6}{0.79} & \cellcolor{gray!6}{0.71} & \cellcolor{gray!6}{0.87}\\
lda > dt & 0.66 & 0.56 & 0.77\\
\bottomrule
\end{tabular}
\end{table}

\hypertarget{convergence-diagnostics-and-execution-times}{%
\subsection{\texorpdfstring{Convergence Diagnostics and execution times \label{sec:conv1}}{Convergence Diagnostics and execution times }}\label{convergence-diagnostics-and-execution-times}}

As previously mentioned, it is important to assess the convergence of an MCMC algorithm with every run. In this study, we utilized Stan \citep{stan229} as the tool to implement the BBT model (\ref{eq:mod1}) and to perform the MCMC. Stan provides various convergence diagnostics data, which are analyzed to determine whether the convergence is acceptable or not. The results of the simplified Stan check are presented below.

\begin{verbatim}
## Checking sampler transitions treedepth.
## Treedepth satisfactory for all transitions.
## 
## Checking sampler transitions for divergences.
## No divergent transitions found.
## 
## Checking E-BFMI - sampler transitions HMC potential energy.
## E-BFMI satisfactory.
## 
## Effective sample size satisfactory.
## 
## Split R-hat values satisfactory all parameters.
## 
## Processing complete, no problems detected.
\end{verbatim}

The MCMC sampling of the model is unproblematic -- in all the examples presented in this paper, including those in later sections, we used 1000 steps of warm-up and 1000 steps of sampling, across 4 chains. The execution time on a modern laptop, such as an Intel i5 running at 1.4 GHz, took no more than 0.5 seconds per chain, with all four chains running simultaneously on the different cores.

\hypertarget{posterior-predictive-check-and-waic}{%
\subsection{\texorpdfstring{Posterior Predictive Check and WAIC \label{sec:ppc1}}{Posterior Predictive Check and WAIC }}\label{posterior-predictive-check-and-waic}}

\begin{figure}
\includegraphics[width=0.7\linewidth]{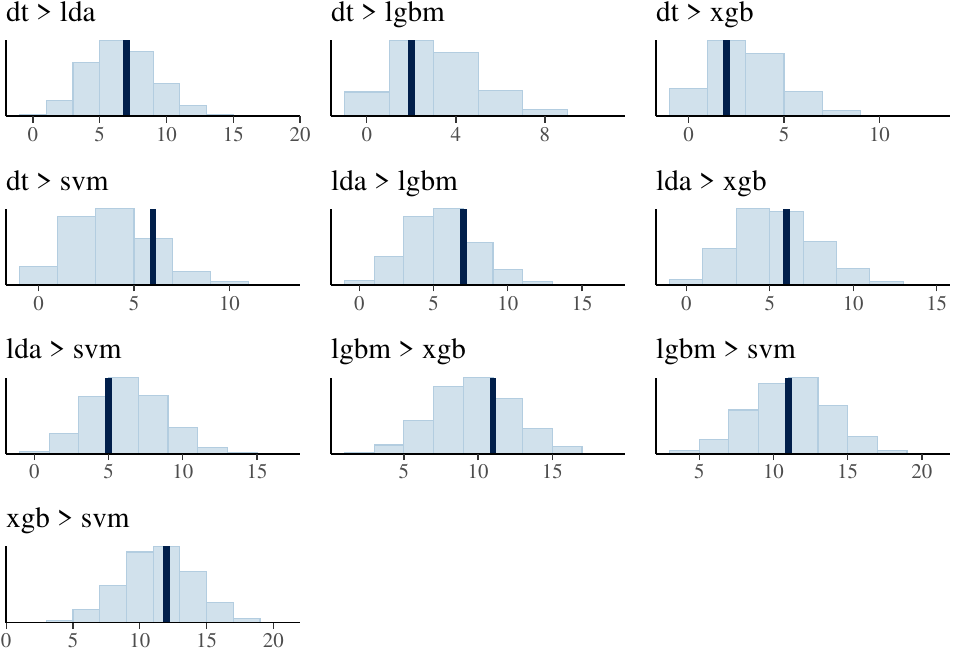} \caption{\label{fig:ppc1}Graphical representation of the PPC. For each observered variable (number of wins of first algoerithm over the second) , the histogram represents the distribution of the generated data, and the wider bar represents the observed number of wins.}\label{fig:xppc1}
\end{figure}

Figure \ref{fig:ppc1} illustrates the results of the posterior predictive check (PCC). The histogram represents the data generated by the Bayesian model, while the vertical bar shows the actual value of the win1 variable from the win/loss table, associated with each pair of algorithms. If the Bayesian model accurately generates the data, the actual values should be centered in the histogram of possible values for that variable.

We also present a non-graphical representation of the PPC by computing the 50\%, 90\%, 95\%, and 100\% HDI (highest density interval) of the generated values for each variable. We then calculate the proportion of the true data that falls within each HDI. Ideally, the proportion of data values that fall within the 90\% HDI should be at least 0.9. Table \ref{tab:ppc1} provides this alternative representation of the PPC.

\begin{table}

\caption{\label{tab:xppc2}\label{tab:ppc1}The table representation of the PPC. Hdi indicates a HDI interval (50\%, 90\%, 95\% and 100\%), and proportion is the proportion of the observed data that falls within the corresponding HDI. }
\centering
\begin{tabular}[t]{rr}
\toprule
\textbf{hdi} & \textbf{proportion}\\
\midrule
\cellcolor{gray!6}{0.50} & \cellcolor{gray!6}{0.8}\\
0.90 & 1.0\\
\cellcolor{gray!6}{0.95} & \cellcolor{gray!6}{1.0}\\
1.00 & 1.0\\
\bottomrule
\end{tabular}
\end{table}

\hypertarget{rope-1}{%
\subsection{\texorpdfstring{ROPE \label{sec:rope2}}{ROPE }}\label{rope-1}}

As discussed above, the Bayesian approach offers the advantage of defining a difference between parameters that may not be meaningful in practical terms, and allows one make statements regarding the likelihood of these parameters being equivalent in a practical sense.

The BBT model provides a simple way to adopt the concept of practical equivalence. The ultimate measure from the BBT model is the probability that a particular algorithm outperforms another. A universal ROPE can be defined for making probability statements, regardless of the metric used to determine superiority between algorithms. We propose that if the probability that one algorithm is better than another falls within the range of 0.45 to 0.55, it can be concluded that the two algorithms are practically equivalent.

This claim is not based on an established community understanding or the author's personal experience with comparing multiple algorithms, but rather on a universal ROPE for probability statements. The choice of the ROPE limits, {[}0.45, 0.55{]}, is somewhat arbitrary, reflecting the author's belief that an algorithm whose probability of being better than another is below 55\% (and above 45\%) is not significantly better than the other. Other researchers may have different intuitions and are free to adjust the ROPE to suit their specific applications.

\begin{figure}
\includegraphics[width=0.7\linewidth]{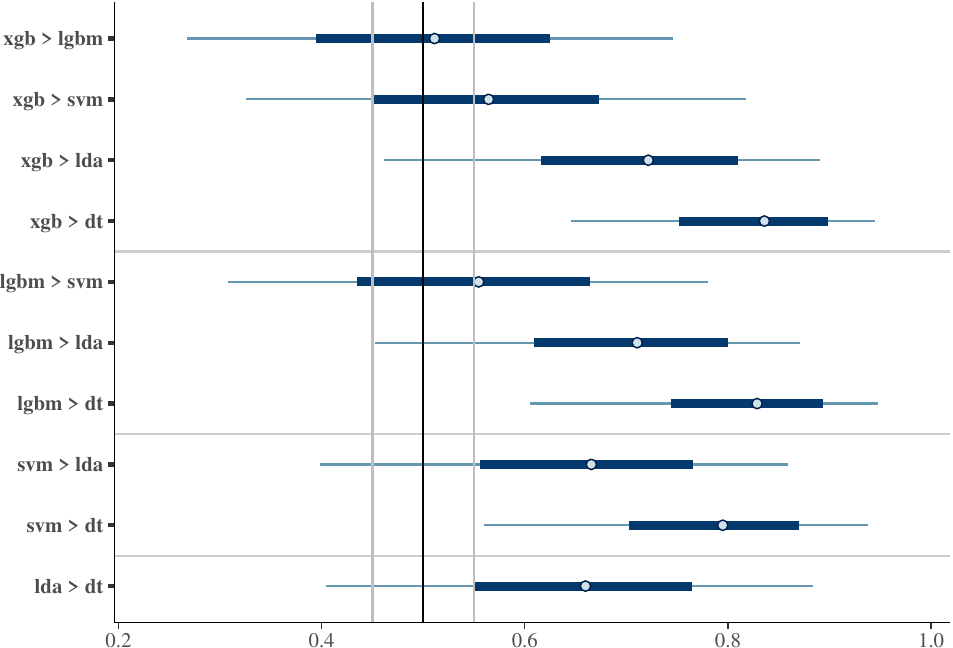} \caption{\label{fig:plot2}Graphical representation of the probability distributions including the ROPE information.}\label{fig:xplot2}
\end{figure}

In Figure \ref{fig:plot2}, the two gray vertical lines represent the lower and upper bounds of the ROPE. This visual representation makes it easy to see if the probability that one algorithm is better than another falls within the ROPE.

Table \ref{tab:main} summarizes the probability distributions, with two additional columns. The {\em in.rope} shows the proportion of samples \(P_s(A \succ\,B)\) that fall within the ROPE interval \([0.45, 0.55]\), The {\em above.50}
displays the proportion of samples \(P_s(A \succ\,B)\) that fall in the interval {[}0.50, 1.00{]}, representing the mass of the probability distribution that algorithm \(A\) is better than algorithm \(B\).

\begin{table}

\caption{\label{tab:xtaball2}\label{tab:main}Table with the summary of the probabilities distributions. Delta is the range of the distribution 89\% HDI; above.50 is the proportion of probabilities samples > 0.50; and in.rope is the proportion of probabilities samples that falls within the ROPE range [0.45 to 0.55].}
\centering
\begin{tabular}[t]{lrrrr}
\toprule
\textbf{pair} & \textbf{mean} & \textbf{delta} & \textbf{above.50} & \textbf{in.rope}\\
\midrule
\cellcolor{gray!6}{xgb > lgbm} & \cellcolor{gray!6}{0.51} & \cellcolor{gray!6}{0.23} & \cellcolor{gray!6}{0.56} & \cellcolor{gray!6}{0.49}\\
xgb > svm & 0.56 & 0.22 & 0.82 & 0.37\\
\cellcolor{gray!6}{xgb > lda} & \cellcolor{gray!6}{0.72} & \cellcolor{gray!6}{0.19} & \cellcolor{gray!6}{1.00} & \cellcolor{gray!6}{0.01}\\
xgb > dt & 0.83 & 0.14 & 1.00 & 0.00\\
\cellcolor{gray!6}{lgbm > svm} & \cellcolor{gray!6}{0.55} & \cellcolor{gray!6}{0.23} & \cellcolor{gray!6}{0.77} & \cellcolor{gray!6}{0.40}\\
\addlinespace
lgbm > lda & 0.71 & 0.19 & 1.00 & 0.01\\
\cellcolor{gray!6}{lgbm > dt} & \cellcolor{gray!6}{0.82} & \cellcolor{gray!6}{0.14} & \cellcolor{gray!6}{1.00} & \cellcolor{gray!6}{0.00}\\
svm > lda & 0.66 & 0.21 & 0.99 & 0.05\\
\cellcolor{gray!6}{svm > dt} & \cellcolor{gray!6}{0.79} & \cellcolor{gray!6}{0.17} & \cellcolor{gray!6}{1.00} & \cellcolor{gray!6}{0.00}\\
lda > dt & 0.66 & 0.21 & 0.99 & 0.05\\
\bottomrule
\end{tabular}
\end{table}

\hypertarget{strong-and-weak-interpretations-of-the-probability-estimates}{%
\subsubsection{\texorpdfstring{Strong and weak interpretations of the probability estimates \label{strong-weak}}{Strong and weak interpretations of the probability estimates }}\label{strong-and-weak-interpretations-of-the-probability-estimates}}

We believe that the four important columns to report are: {\em mean},
{\em delta} (the difference between the {\em high} and {\em low} values of the HDI)
{\em in.rope}, and {\em above.50}. In particular, the {\em mean} and the {\em above.50}
measures measures play a crucial role in what we refer to as the \textbf{strong} and the
\textbf{weak interpretations} of the probability estimates. The
BBT model generates a set of numbers \(P_s(A \succ\,B)\) which we interpreted as probabilities that algorithm \(A\) is better than algorithm \(B\). And in fact,
these numbers are used in the BBT model as the parameters of the
binomial distribution that are interpreted as probabilities of the
event happening.

Under the strong interpretation, we understand each of \(P_s(A \succ\,B)\)
as \emph{probabilities} that \(A\) is better than \(B\) in the sense that
in the long run, for a large number of data
sets, the proportion of times \(A\) wins from \(B\) should approach that
number. In the strong interpretation, the {\em mean} column is the best estimation of how much better algorithm \(A\) is compared to algorithm \(B\). The {\em delta} column or both
{\em low} and {\em high} are estimates of the uncertainty surrounding that probability.

The weak interpretation views each \(P_s(A \succ\,B)\) as a \emph{measure of the superiority} of A over B, expressed as a number ranging from 0.0 to 1.0. A value less than 0.5 indicates that B is better than A. Under this interpretation, \(P_s(A \succ\,B)\) represents evidence in favor of A's superiority over B, rather than a guarantee of future outcomes. The value of {\em above.50} reflects the degree of confidence one can have in the superiority of A over B. For example, if 90\% of the evidence (\(P_k(A \succ\,B)\)) is above 0.5, one can have 90\% confidence that A is better than B.''

The {\em in.rope} measure combines elements of both interpretations. While it calculates the proportion of evidence that falls within a specific interval (from 0.45 to 0.55), this range was determined based on the strong interpretation's perspective.

There are two reasons for presenting these two interpretations rather than solely relying on the strong interpretation. As it will be discussed in Section \ref{compare-tests}, the weak interpretation is necessary when comparing the BBT model to previous frequentist models. Also, as will be discussed in Section \ref{predictive}, the BBT model may not be well-calibrated for making predictions about future algorithm results under the strong interpretation, but it is more accurate when evaluated using the weak interpretation.

\hypertarget{further-results-missing-data-comparisons-against-a-control-and-too-many-comparisions}{%
\subsection{\texorpdfstring{Further results: missing data, comparisons against a control, and too many comparisions \label{sec:xextra}}{Further results: missing data, comparisons against a control, and too many comparisions }}\label{further-results-missing-data-comparisons-against-a-control-and-too-many-comparisions}}

Regarding missing values, the cases where an algorithm cannot run on
one or more data sets, the BBT model simply does not count it as a win or a
loss for that algorithm in comparison to the others. For example, let us assume that the algorithm \emph{xgb} does not run
on the first two data sets in Table \ref{tab:sstab1} (data sets
\texttt{biomed} and \texttt{breast}). The resulting win/loss table is displayed in Table
\ref{tab:miss22a}, which should be contrasted with the win/loss table in
Table \ref{tab:wintab2}, and the summary results are displayed in
Table \ref{tab:miss22b}, which should be contrasted with the results
in Table \ref{tab:main}.

\begin{table}[ht]
\centering
\subfloat[The win/loss table]{\label{tab:miss22a}\scalebox{1.0}{\input{./miss1}}}\quad
\subfloat[Results for the corresponding BBT model]{\label{tab:miss22b}\scalebox{1.0}{\input{./miss2}}}
\caption{Results when *xrg* does not run on the first two data sets.}
\label{tab:miss22}
\end{table}

It is believed that Bayesian tests do not suffer from an decrease in the power to detect (statistically significant) differences between the algorithms as the number of algorithms increases. Therefore, there is no need to distinguish comparison against a control and all pairwise comparisons. If one is interested in displaying just the comparisons of the control algorithm against its competitors, one only needs to limit the rows of the summary table that are shown.

The insensibility to the number of algorithms being compared has implications when comparing a large number of algorithms. As discussed above, the literature comparing 50 to 100 algorithms avoid statistical tests altogether. An alternative is to perform a two step procedure. If the algorithms can be naturally grouped into (few) families, one compares the algorithms within a single family, to select the best of that family, and then compare the ``best representatives'' of each family among each other (using the full frequentist tests). This was done, for example within the context of comparing imbalanced data algorithms by \citet{lopez2013insight}.

For BBT, under the strong interpretation, there is no need to perform the two steps procedure; there is no large and biased difference
between the probability estimates when comparing a large number of algorithms and a small one. Figure \ref{fig:fewmany2}
displays the results of comparing the mean probability \(P_s(A \succ\,B)\) of a random sample of three
algorithms when all the 16 algorithms' results on a random sample of 20 data sets from the $\ell\ell$-results are fed to the BBT procedure, contrasted to the mean when only the results from those 3 algorithms are fed to BBT. This procedure is repeated 40 times. The difference between the two mean estimates has mean 0.006, median 0.010, 1st quartile -0.020, and 3th quartile 0.030. That is, although the new mean probability is not necessarily the same as when tested for all algorithms, the difference in magnitude is small, and there is no bias - the difference is as likely to be positive as it is to be negative.

\begin{figure}
\includegraphics[width=0.7\linewidth]{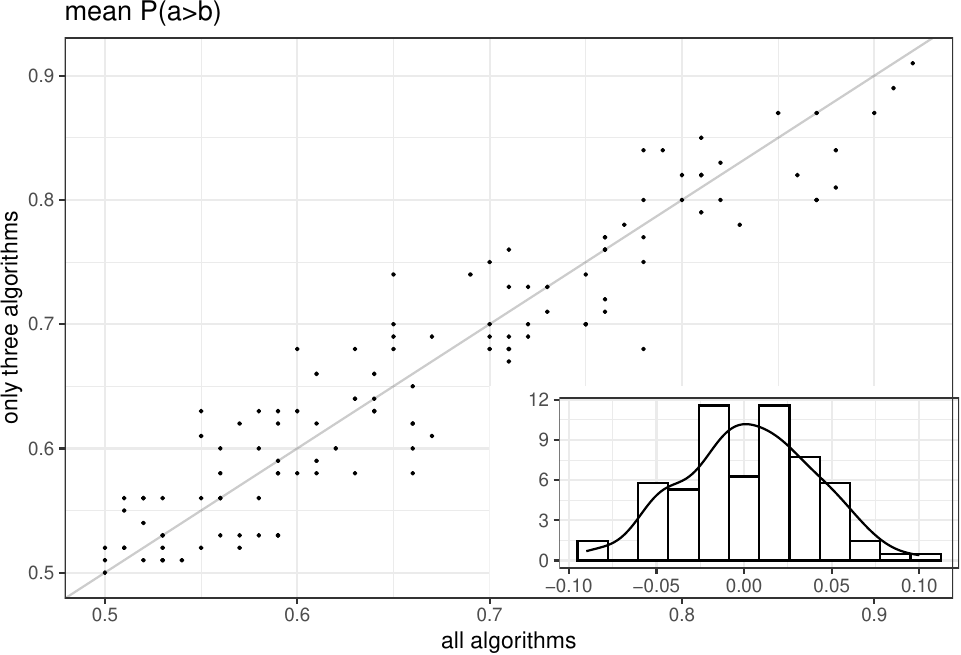} \caption{\label{fig:fewmany2}The results for mean P(a>b) of comparing 3 algorithms when all 16 algorithms' results are fed to BBT (on a random sample of 20 data sets) and when only the results for the three algorithms are fed to the BBT procedure. The inset is the distribution of the differences between the two avergage probability of winning.}\label{fig:fewmany2}
\end{figure}

Unfortunately, the insensibility to large number of comparisons is not true for the weak interpretation. Figure \ref{fig:fewmany3} compare the {\em above.50} results for the full 16 algortithms comparison and for a limited 3 algorithm comparison. There is a clear bias in the limited number of comparisons but the direction of the bias is surprising. The {\em above.50} numbers when only comparing 3 algorithms are \emph{smaller} than the corresponding numbers when the full set of algorithms, which indicates that with more algorithms being compared the procedure will be \emph{more} sure of the difference between them. That is in the opposite direction one would expect from the frequentist tests: many more algorithms will decrease the power of the test and reduce the number of pairs which will be classified as statistically significant.

\begin{figure}
\includegraphics[width=0.7\linewidth]{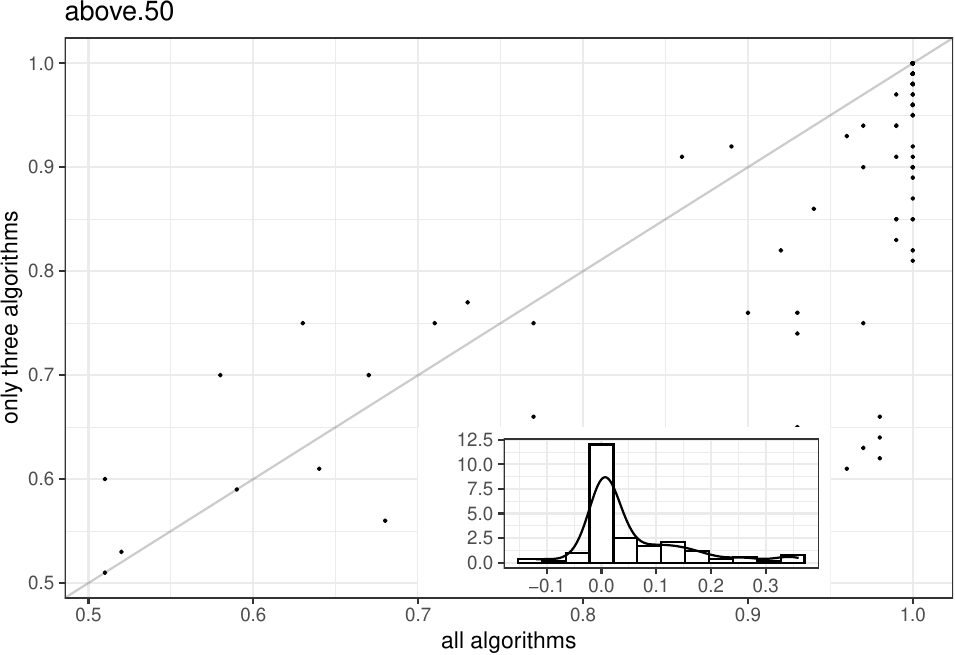} \caption{\label{fig:fewmany3}The results for above.50 of comparing 3 algorithms when all 16 algorithms' results are fed to BBT (on a random sample of 20 data sets) and when only the results for the three algorithms are fed to the BBT procedure. The inset is the distribution of the differences between the two avergage probability of winning.}\label{fig:fewmany3}
\end{figure}

\hypertarget{what-counts-as-a-win-folds-and-local-rope}{%
\section{\texorpdfstring{What counts as a win? Folds and local rope \label{sec:lrope}}{What counts as a win? Folds and local rope }}\label{what-counts-as-a-win-folds-and-local-rope}}

Typically, the final performance measure for a particular algorithm for
a particular data set is obtained by averaging the results from some form of
repeated cross-validation, where the algorithm is trained on different
subsets of the data set and its performance is measured on the
corresponding test subsets. Standard forms of repeated
cross-validations are k-fold, repeated k-folds, repeated train/test
split, and bootstrapped samples of the data set. In each case, the data set is divided into \(k\) pairs of subsets, \(TR_i\) (train) and \(TE_i\) (test) such that
\(TR_i \cup TE_i = DS\) and \(TR_i \cap TE_i = \emptyset\) (\(DS\) is the
whole data set). The term \textbf{fold} will be used to refer to each \(TE_i\) as a \textbf{fold},
although we do not assume that k-fold cross-validation is being
used -- almost all cross-validation procedures can be used.

For the data in this research, the mean of a 4-fold cross-validation was calculated for each algorithm on each data set. Additionally, the folds were fixed and identical for all algorithms, meaning that for the first fold, all algorithms were trained on \(TR_1\) and tested on \(TE_1\), and so on.

Let us consider the entries for the ``cmc'' data set from Table \ref{tab:sstab1}, repeated in Table \ref{tab:small1}. The entries for the lgbm and xgb algorithms are 0.525 and 0.524, respectively. Although the difference between these two values is small, it still counts as a win for lgbm, just as the much larger difference in accuracy between dt and lgbm also counts as a win.

\begin{table}

\caption{\label{tab:xsmall1}\label{tab:small1}Detail of the base results for the cmc data set}
\centering
\begin{tabular}[t]{lrrrrr}
\toprule
\textbf{db} & \textbf{dt} & \textbf{lda} & \textbf{lgbm} & \textbf{xgb} & \textbf{svm}\\
\midrule
cmc & 0.455 & 0.513 & 0.525 & 0.524 & 0.544\\
\bottomrule
\end{tabular}
\end{table}

If we examine the performance of the algorithms on each fold separately, as displayed in Table \ref{tab:small2} for the ``cmc'' data set, the small difference in accuracy between lgbm and xgb becomes even less convincing as a win for lgbm. In this case, since we used the same fold 1, fold 2, etc. for all algorithms, it is reasonable to compare the performance of each algorithm on each fold. In this scenario, lgbm wins on two of the four folds, but loses on the other two. As a result, if we use the individual folds as evidence instead of the mean of their results, lgbm would receive two wins and xgb would receive two wins, rather than a single win for lgbm based on the mean accuracy.

From another perspective, if we take into account the standard deviations of the measures on the folds for each algorithm, as shown in Table \ref{tab:small2}, the difference in the means of both algorithms (0.001) is much smaller compared to the standard deviations (0.02 and 0.03). In some intuitive sense, the difference in the averages that led to the win for lgbm is much smaller than the ``noise level'' of the evaluation procedure itself, given the variability of the measures within each classifier in the folds.

This leads us to two ways of interpreting the results: one that takes into account each fold as individual sources of evidence to determine the wins and losses of the algorithms, and another that considers the difference between the means across all folds while also taking into account the ``noise level'' derived from the variability among the folds. Both methods aim to reduce the strength of evidence that lgbm wins over xgb, resulting in a tie between the two algorithms.

In this research, we will adopt the latter approach, which considers the ``noise level.'' We argue that a difference of 0.001 in the means, given that the variability of the measures in the folds for each classifier is at least 10 times higher, should not be considered a win, but rather a tie between the two algorithms. We refer to this approach as the \textbf{local ROPE}, which is a threshold below which differences between two classifiers are considered unimportant. However, as we will see, the local ROPE is not a fixed number but it depends on the results of the two algorithms on the different folds.

In regards to the first line of reasoning, where the folds are used as the source of evidence for wins and losses, we believe that issues such as the dependence of the fold results on each other would make the analysis too complex. This conclusion was also reached by \citet{benavoli2017time} in their analysis, and as such, we will leave this approach for further exploration in future research.

\begin{table}

\caption{\label{tab:xsmall2}\label{tab:small2}Detail of the base results for the cmc data set}
\centering
\begin{tabular}[t]{llrrr}
\toprule
\textbf{db} & \textbf{fold} & \textbf{lgbm} & \textbf{xgb} & \textbf{diff}\\
\midrule
cmc & 1 & 0.547 & 0.531 & 0.016\\
cmc & 2 & 0.522 & 0.538 & -0.016\\
cmc & 3 & 0.503 & 0.481 & 0.022\\
cmc & 4 & 0.527 & 0.546 & -0.019\\
sd &  & 0.018 & 0.029 & 0.001\\
\bottomrule
\end{tabular}
\end{table}

\hypertarget{local-rope}{%
\subsection{\texorpdfstring{Local ROPE \label{localrope}}{Local ROPE }}\label{local-rope}}

In almost all statistical tests, one has two sets of measures, and the goal is
determine whether there is enough evidence that
the difference of the means or some other summary measure of the two sets is ``real'' or not. This
is exactly the problem in hand: should one consider the difference of
the means of the folds as ``real'' -- and thus that one algorithm wins over the other -- or not?

Cohen's D is a measure of effect size between two sets of data. If the two sets have the same number of data, as it is our
case, the Cohen D is computed as the difference between the means,
divided by an ``average'' standard deviation of the two sets, where the
``average'' standard deviation is actually the square root of the average
variance of the two sets. This is displayed in Equation \ref{eq:d} where \(\mu_1\) and \(\sigma_{1}\) are the mean and standard deviation of the fold measures for the first algorithm, and similarly \(\mu_2\) and \(\sigma_{2}\) for the second algorithm.

Cohen D is the measure of the separation between the the means of two sets of measures as a
proportion of the standard deviation, and can be seen as a signal-to-noise
ratio measure: the difference in means is the signal, and the
``average'' standard deviation is the noise.

\begin{equation} \label{eq:d}
d = \frac{\mu_1 - \mu_2}{\sqrt{\frac{\sigma_1^2 + \sigma_2^2}{2}}} 
\end{equation}

We can compute the Cohen D of two sets of fold measures, and consider
that there is no important difference, and thus, a tie between the two
algorithms, if the D is below a threshold \(d_{\min}\) which we will
call the \textbf{local ROPE threshold}. Therefore, if:

\begin{equation} \label{eq:dmin}
\mu_1 - \mu_2 \le d_{\min} \sqrt{\frac{\sigma_1^2 + \sigma_2^2}{2}} 
\end{equation}
we should consider that there was a tie between algorithms 1 and 2 for that data set.

We will argue that the threshold can be safely set to the value of 0.4
using the theory of power analysis for t-tests. Type 1 and type 2 errors in statistical test are a false positive (claiming that there is a difference when there is no difference) and a false negative error (claiming that there is no difference when there is one), and their probabilities are indicated by \(\alpha\) and \(\beta\).
The power analysis relates \(\alpha\), \(\beta\), the effect
size of the measure, and the number of samples in each set.
Unfortunately, the relation between these variables is almost never
displayed as an equation, but as tables \citep[ch.~2]{cohen1988statistical}
or embedded into programs, such as G*power \citep{faul2007g} or
the \texttt{pwr} R package \citep{pwrR}. We will show the results of running the
\texttt{pwr} package.

For our present goals, there is no conceptual difference
between false positive and false negative errors. We want to find out
whether the two sets of fold measures indicate that the difference
between the means is ``real'' or ``not real'', and erring to one side is
not worse than erring to the other. Thus, let us assume a 30\% probability of
making a mistake, both false positive or false negative, that is, \(\alpha = 0.3\) and \(\beta = 0.3\). Assuming a Cohen D of 0.4, the necessary number of data
in each set is given by running the \texttt{pwr} function. In that function \texttt{sig.level} is \(\alpha\), \texttt{power}is \(1-\beta\), and \texttt{d} is Cohen's D.

\begin{quote}
\begin{verbatim}
 pwr.t.test(n = NULL, d = 0.4, sig.level = 0.3, power = 0.7)

 Two-sample t test power calculation 
           n = 30.18637
           d = 0.4
   sig.level = 0.3
       power = 0.7
 alternative = two.sided
 NOTE: n is number in *each* group
\end{verbatim}
\end{quote}

That is, one would need at least 30 measures in each set to be able
to find a true difference or a true non-difference with 70\%
probability. But the traditional cross-validations in machine
learning are from 3 to 10 folds. That is, using the usual cross-validation
in machine learning, a minimum effect size of 0.4 is very
safe - one would not be able to detect differences whose effect sizes
are 0.4 or below, if one requires a 70\% of sensitivity and
specificity. If one is using 10 repetitions of 10-folds as cross-validation,
one can use \(d_{\min} = 0.2\).

The discussion above assumes that the two samples of fold measures are
not paired, that is, that possibly different
folds were used in the evaluation of the different algorithms. But if
the researchers have control over it, they can use the same folds for
all algorithms. For the paired case, the definition of
Cohen's D is somewhat different than the one presented in
Equation \ref{eq:d}. Instead of dealing with the mean and standard deviations
of the two sets, one should compute the mean and standard deviation of
the differences between the corresponding paired data in the two sets. In Equation \ref{eq:dz}, \(\mu{X_1 - X_2}\) is the mean and \(\sigma_{X_1 - X_2}\) the standard deviation of the pairwise differences of the corresponding folds for algorithm 1 and 2,

\begin{equation} \label{eq:dz}
d_z = \frac{\mu_{X_1 - X_2}}{\sigma_{X_1 - X_2}} = \frac{\mu_1 - \mu_2}{\sigma_{X_1 - X_2}} 
\end{equation}

The power analysis for paired samples is also somewhat different, and
with the same numbers as before (\(\alpha = 0.3\) and \(\beta = 0.3\)),
and using Equation \ref{eq:dz} for the effect size calculation, the
resulting lower bound for the number of samples is 15, lower than the
case for unrelated samples, but still well above the usual number of
folds used in machine learning evaluations.

\begin{quote}
\begin{verbatim}
  pwr.t.test(n = NULL, d = 0.4, sig.level = 0.3, power = 0.7, type="paired")

  Paired t test power calculation 
           n = 15.53464
           d = 0.4
   sig.level = 0.3
       power = 0.7
 alternative = two.sided
 NOTE: n is number of *pairs*
\end{verbatim}
\end{quote}

The same decision process as described in \ref{eq:dmin} can be
followed, using the same \(d_{min}\) threshold of 0.4, but using the
paired definition for the effect size.

\begin{equation} \label{eq:dmin2}
\mu_1 - \mu_2 \le d_{\min} \sigma_{X_1 - X_2}
\end{equation}

\hypertarget{how-to-deal-with-ties}{%
\subsection{\texorpdfstring{How to deal with ties? \label{test-ties}}{How to deal with ties? }}\label{how-to-deal-with-ties}}

The local ROPE concept introduces new ties to the win/loss table, as it is designed to do.
The standard ways of dealing with ties in the Bradley-Terry model are:

\begin{itemize}
\tightlist
\item
  add: add the ties as victories to both players involved.
\item
  spread: add the ties as half a victory to each player involved
\item
  forget: do not add ties as victories to any of the players.
\end{itemize}

Another alternative is to use an extension of the Bradley-Terry model
that includes ties, for example, the one proposed by
\citet{davidson1970extending}. The Davidson model is displayed in Equation
\ref{eq:david} and it includes a new parameter \(\nu\), similar to the
\(\beta_i\). \(\nu\) controls how likely are ties ``in that sport'', despite the
differences between the players. If \(\nu \rightarrow -\infty\), the probability
of a tie between player \(i\) and player \(j\) will be 0, meaning there are no ties;
if \(\nu \rightarrow \infty\), \(P( i \mbox{~ties~} j)\) will be 1, regardless of
the players' different \(\beta\). Finally, for \(\nu = 0\), and if \(\beta_i = \beta_j\) then the probability of a tie is \(1/3\).

\begin{align}
 P(i \succ\,j | \mbox{~no tie~}) &= \frac{\exp \beta_i}{\exp \beta_i +\exp \beta_j + \exp(\nu + (\beta_i + \beta_j)/2) }  \label{eq:david}\\
P( i \mbox{~ties~} j) &= \frac{\exp (\nu + (\beta_i + \beta_j)/2) }{\exp \beta_i +\exp \beta_j + \exp (\nu + (\beta_i + \beta_j)/2 ) } \nonumber
\end{align}

We will compare the various policies for dealing with ties, using a repeated
experiment as described above. To evaluate how well each policy fits the actual
data, we will use the posterior predictive check and the WAIC. With respect to
the WAIC, while the numerical value itself can be difficult to interpret, when
comparing two models, a lower WAIC value indicates a better fit.

Tables \ref{tab:tiesppc1} and \ref{tab:tiesppc1b} present the average results of the WAIC and PPC for the repeated experiments comparing various methods for handling ties. These results are based on averaging across the ss, mm, sl use cases, and the $\ell\ell$-results, taking into account whether the local ROPE or the paired local ROPE was used. The results clearly demonstrate that the Davidson model is significantly inferior to the others in terms of both WAIC and PPC. The add, forget, and spread policies are all equivalent, and for the purpose of this paper, we have arbitrarily chosen to use the spread policy.

The poor performance of the Davidson model is unexpected, given that it was specifically designed to handle ties, while the other policies are \emph{ad hoc} in nature. Table \ref{tab:tiesppc2} further illustrates this point, as the PPC summary shows that the wins and ties are not well-calibrated according to their corresponding HDI.

\begin{table}

\caption{\label{tab:tiesxz1}\label{tab:tiesppc1}The PPC of ss and mm use cases for the different policies. WAIC is the mean WAIC result; h50 is the proportion of observed valies that fall in the 50\% HDI; h90, h95, and h100 are the proportions for the 90\%, 95\% and 100\% HDIs.}
\centering
\begin{tabular}[t]{lrrrrrrrrrr}
\toprule
\multicolumn{1}{c}{ } & \multicolumn{5}{c}{ss} & \multicolumn{5}{c}{mm} \\
\cmidrule(l{3pt}r{3pt}){2-6} \cmidrule(l{3pt}r{3pt}){7-11}
\textbf{\textbf{policy}} & \textbf{\textbf{waic}} & \textbf{\textbf{h50}} & \textbf{\textbf{h90}} & \textbf{\textbf{h95}} & \textbf{\textbf{h100}} & \textbf{\textbf{waic}} & \textbf{\textbf{h50}} & \textbf{\textbf{h90}} & \textbf{\textbf{h95}} & \textbf{\textbf{h100}}\\
\midrule
add & 42.41 & 0.87 & 1.00 & 1.00 & 1 & 225.51 & 0.73 & 1.00 & 1.00 & 1.00\\
davidson & 118.59 & 0.40 & 0.78 & 0.86 & 1 & 873.01 & 0.29 & 0.58 & 0.66 & 0.88\\
forget & 40.27 & 0.78 & 0.99 & 1.00 & 1 & 227.61 & 0.61 & 0.96 & 0.99 & 1.00\\
spread & 41.41 & 0.82 & 1.00 & 1.00 & 1 & 223.76 & 0.68 & 0.99 & 1.00 & 1.00\\
\bottomrule
\end{tabular}
\end{table}

\begin{table}

\caption{\label{tab:tiesxz2}\label{tab:tiesppc1b}The PPC of sl and ll use cases the different policies. WAIC is the mean WAIC result; h50 is the proportion of observed valies that fall in the 50\% HDI; h90, h95, and h100 are the proportions for the 90\%, 95\% and 100\% HDIs.}
\centering
\begin{tabular}[t]{lrrrrrrrrrr}
\toprule
\multicolumn{1}{c}{ } & \multicolumn{5}{c}{sl} & \multicolumn{5}{c}{ll} \\
\cmidrule(l{3pt}r{3pt}){2-6} \cmidrule(l{3pt}r{3pt}){7-11}
\textbf{\textbf{policy}} & \textbf{\textbf{waic}} & \textbf{\textbf{h50}} & \textbf{\textbf{h90}} & \textbf{\textbf{h95}} & \textbf{\textbf{h100}} & \textbf{\textbf{waic}} & \textbf{\textbf{h50}} & \textbf{\textbf{h90}} & \textbf{\textbf{h95}} & \textbf{\textbf{h100}}\\
\midrule
add & 61.62 & 0.81 & 0.97 & 0.99 & 1.00 & 760.49 & 0.57 & 0.95 & 0.97 & 1.00\\
davidson & 325.65 & 0.21 & 0.51 & 0.56 & 0.78 & 4730.36 & 0.17 & 0.38 & 0.43 & 0.70\\
forget & 65.46 & 0.67 & 0.95 & 0.97 & 0.99 & 823.52 & 0.39 & 0.85 & 0.91 & 0.99\\
spread & 62.25 & 0.75 & 0.97 & 0.99 & 1.00 & 773.98 & 0.47 & 0.92 & 0.95 & 1.00\\
\bottomrule
\end{tabular}
\end{table}

\begin{table}

\caption{\label{tab:ties2}\label{tab:tiesppc2} The Davidson model on the large-large results. Ties is the proportion of ties that fall within the corresponding HDI for the ties generated data.}
\centering
\begin{tabular}[t]{rrr}
\toprule
\textbf{hdi} & \textbf{proportion} & \textbf{ties}\\
\midrule
0.50 & 0.21 & 0.27\\
0.90 & 0.47 & 0.43\\
0.95 & 0.52 & 0.48\\
1.00 & 0.92 & 0.78\\
\bottomrule
\end{tabular}
\end{table}

\hypertarget{comparison-with-standard-approaches}{%
\section{\texorpdfstring{Comparison with standard approaches \label{compare-tests}}{Comparison with standard approaches }}\label{comparison-with-standard-approaches}}

Let us compare the results of using the BBT procedure with that of using some of the standard comparison procedures. Section \ref{sec:comp-demsar} compare the BBT with Demsar's procedure on a different set of use-cases; section \ref{sec:comp-wil} compares with the pairwise Wilcoxon procedure, and section \ref{sec:comp-bsr} compares with the Bayesian signed rank procedure.

\hypertarget{demsars}{%
\subsection{\texorpdfstring{Demsar's \label{sec:comp-demsar}}{Demsar's }}\label{demsars}}

\begin{figure}
\begin{floatrow}
\ffigbox{%
  \includegraphics[scale=0.5]{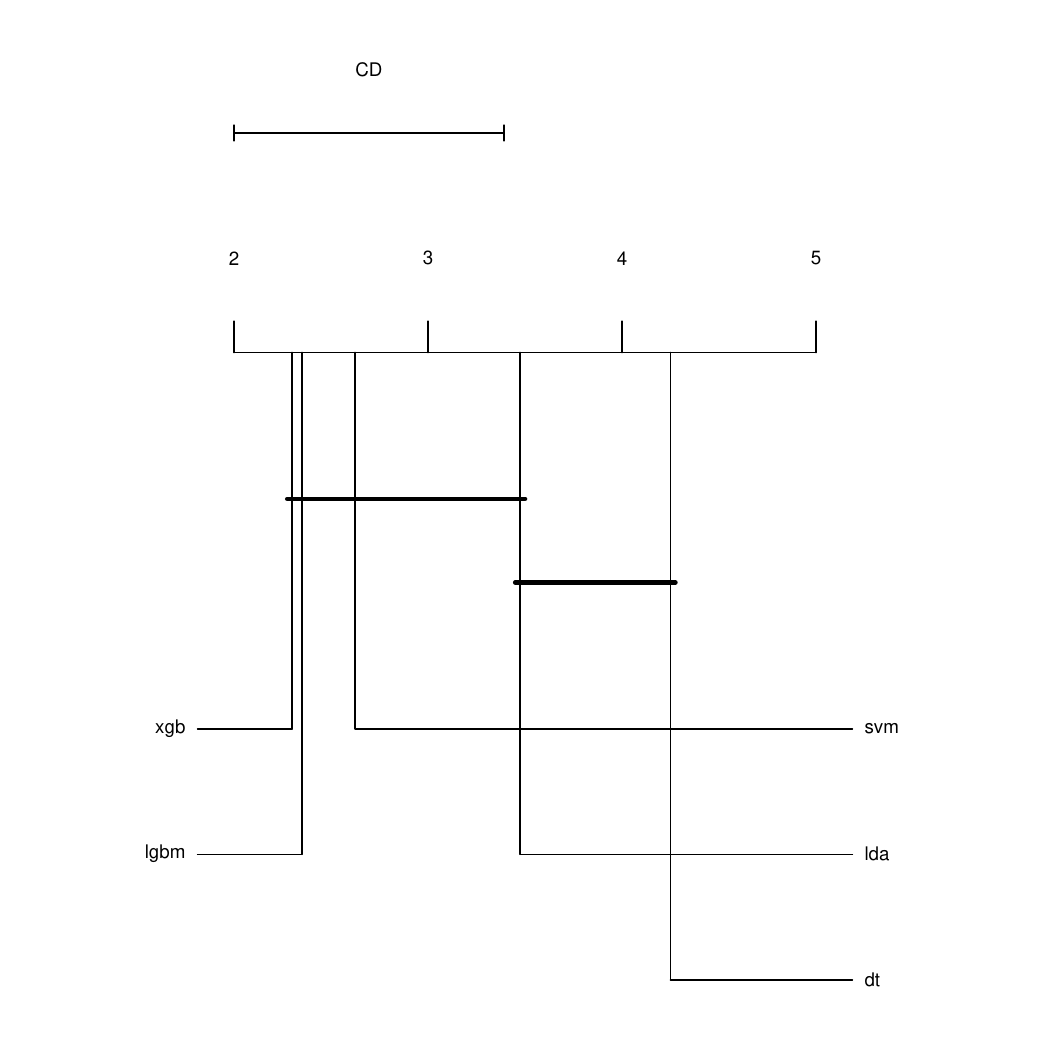}%
}{%
  \caption{The CD plot of the base results}\label{fig:demsar1}.%
}
\capbtabbox{%
\input{./tdemsar1}
}{%
  \caption{The table representation of the CD plot}\label{tab:demsar22}%
}
\end{floatrow}
\end{figure}

The Demsar's procedure starts with the evaluation of the Friedman test, which results in a p-value \(\le 0.05\) for the base results.
The next step is the Nemenyi test, which results in the Critical Difference plot displayed in Figure \ref{fig:demsar1}.
The results of significant and non-significant differences can be displayed in a tabular form as in
Table \ref{tab:demsar22}.

We refer the reader to Table \ref{tab:main} for the BBT results regarding the base results. Here, we are faced with the strong vs weak interpretations of the probability estimates. Under the \emph{strong} interpretation, it would be reasonable to make an equivalence between a frequentist test claiming statistically significant difference between algorithm A and B (with 95\% confidence) and that the mean estimate that \(P(A \succ\,B) \ge 0.95\). These two concepts are not the same, but it is reasonable to make the equivalence to compare them. In this case, none of the claims of superiority made by the BBT model reaches the level of 95\%. The highest mean is 0.84 (\emph{xgb} against \emph{dt}).

But under the weak interpretation, using {\em above.50} \(\ge 0.95\) as the equivalence of statistical significance, the BBT procedure finds 5 differences that one would call significant, as opposed to three found by Demsar's procedure. In the base results, the differences between \emph{lgbm} and \emph{lda}, and \emph{xgb} and \emph{lda} are detected as ``significant'' by BBT, but they were not detected as such by the Demsar's procedure.

We ran the repeated experiments to verify how frequent are these results when comparing the BBT model with the traditional frequentist approach proposed by Demsar: how many times does BBT find a significant difference (in the sense that the {\em above.50} result is larger than 0.95) where Demsar does not find it significant, how many times the reverse happens, and how many times the aggregated ranking computed by one method did agree with the one computed by the other method.

The results are:

\begin{itemize}
\item
  \emph{ss:} For the 100 pairs of comparisons (10 from each ss), the BBT
  method classified 40 of them as significantly different, when
  Demsar did not determine them to be. No pairs were found
  significantly different by Demsar's procedure and not by BBT. Finally,
  only 3 of the 10 aggregated rankings were found to be the same.
\item
  \emph{mm:} For the 450 pairs, BBT found 187 that were not significant
  according to the Demsar procedure, none the other way, and 5 of
  the 10 aggregated rankings were the same.
\item
  \emph{sl:} For the 50 pairs, 11 were found as significant only by BBT
  and none the other way, and 5 aggregated rankings were in
  agreement.
\item
  \emph{$\ell\ell$ results:} For 120 pairs of comparisons, BBT found 41 that were
  not significant by Demsar's, none the other way, and the
  aggregated ranking did not agree.
\end{itemize}

Thus there is strong evidence that the BBT procedure is stronger, in
the sense that it finds more significant differences than Demsar's
procedure, and that it supersedes Demsar's procedure, in the sense
that it does not miss any significant difference determined by
Demsar's. Furthermore, usually 50\% of the time, the two procedures do
not fully agree on the final aggregated ranking.

For all but the $\ell\ell$-results, there was no case in which {\em in.rope}
\(\ge 0.95\). For the $\ell\ell$ results, there were 8 cases in which
Demsar's procedure sees non-significant differences but the BBT can
make a stronger claim that the two algorithms are equivalent. There is
an interesting case, which did not occur with our data, but
\citet{benavoli2017time} report, of two algorithms whose difference is
significant but are practically equivalent. The frequentist approach
has enough evidence to claim that the algorithms are different, but
the Bayesian approach claims that the difference does not matter! But
this is expected: with enough data all algorithms will be found
statistically different from each other, even those whose difference
is irrelevant.

\hypertarget{pairwise-wilcoxon}{%
\subsection{\texorpdfstring{Pairwise Wilcoxon \label{sec:comp-wil}}{Pairwise Wilcoxon }}\label{pairwise-wilcoxon}}

Table \ref{tab:wilcox22} reports the result of the pairwise Wilcoxon
tests, using the Horchberg p-value adjustment procedure. The column
{\em p-value} indicates the adjusted p-value of the comparison, and as
usual, the comparisons where p.value \(\le 0.05\) are considered
significant.

\begin{table}[ht]
\centering
\subfloat[Pairwise Wilcoxon: Order of the algorithms]{\scalebox{1.0}{\input{./twilc2}}}\quad
\subfloat[Pairwise Wilcoxon: Significance of differences displayed as a table.]{\label{tab:pairwil1}\scalebox{1.0}{\input{./twilc1}}}
\caption{Pairwise Wilcoxon tests - order of the algorithms and significance of differences displayed as a table."}
\label{tab:wilcox22}
\end{table}

Following the same procedure described above of considering
{\em above.50} \(\ge 0.95\) as ``equivalent'' to p.value \(\le 0.05\), we
see that the pairwise Wilcoxon tests only detect two significant
differences, for the five detected by the BBT model. Also, the
aggregated rank is not the same for the pairwise Wilcoxon and BBT. The
second and third best algorithms and the fourth and fifth (according
to BBT) are in reverse order for the pairwise Wilcoxon procedure,
although it does not find these two pairs of algorithms as
significantly different from each other.

We ran the replication experiments as above and
the results are:

\begin{itemize}
\item
  \emph{ss:} For the 100 pairs the BBT found 21 beyond the pairwise
  Wilcoxon, only one pair was detected by the pairwise Wilcoxon and
  missed by the BBT, and 2 of the 10 aggregated rankings were the
  same.
\item
  \emph{mm:} For the 450 pairs, BBT found 126 beyond the pairwise
  Wilcoxon, 6 missed, and no aggregated ranking was the same.
\item
  \emph{sl:} For the 50 pairs, BBT found 6 beyond the pairwise Wilcoxon,
  1 missed, and only one aggregated ranking was the same.
\item
  \emph{$\ell\ell$ results:} For 120 pairs of comparisons, BBT found 21 beyond
  the pairwise Wilcoxon, 1 missed, and the methods did not agree on
  the aggregated ranking.
\end{itemize}

BBT seems to be stronger than the pairwise Wilcoxon but not all pairs
deemed significant by the pairwise Wilcoxon are found by BBT. The two
procedures fully agree on very few of the aggregated rankings. Also,
for the $\ell\ell$-results, there were 7 cases in which the pairwise Wilcoxon
procedure sees non-significant difference but the BBT can make a
stronger claim that the two algorithms are equivalent.

\hypertarget{pairwise-bayesian-signed-rank}{%
\subsection{\texorpdfstring{Pairwise Bayesian signed rank \label{sec:comp-bsr}}{Pairwise Bayesian signed rank }}\label{pairwise-bayesian-signed-rank}}

Following the caveat that it is unclear whether one should use the
Bayesian signed rank (BSR) on multiple comparisons, Table
\ref{tab:pairbays1} displays the results of the
comparisons. \citet{benavoli2017time} suggest that one should report the
probabilities that the difference falls within the 0.1 ROPE
({\em in.rope}) and the probability that it falls above the ROPE
({\em above.rope}). We also report the probability that the
probability is above 0 ({\em above.0}), that is, that the difference
between the two mean parameter values is positive, that is, that one
algorithm is better than the other.

\begin{figure}
\includegraphics[width=0.9\linewidth]{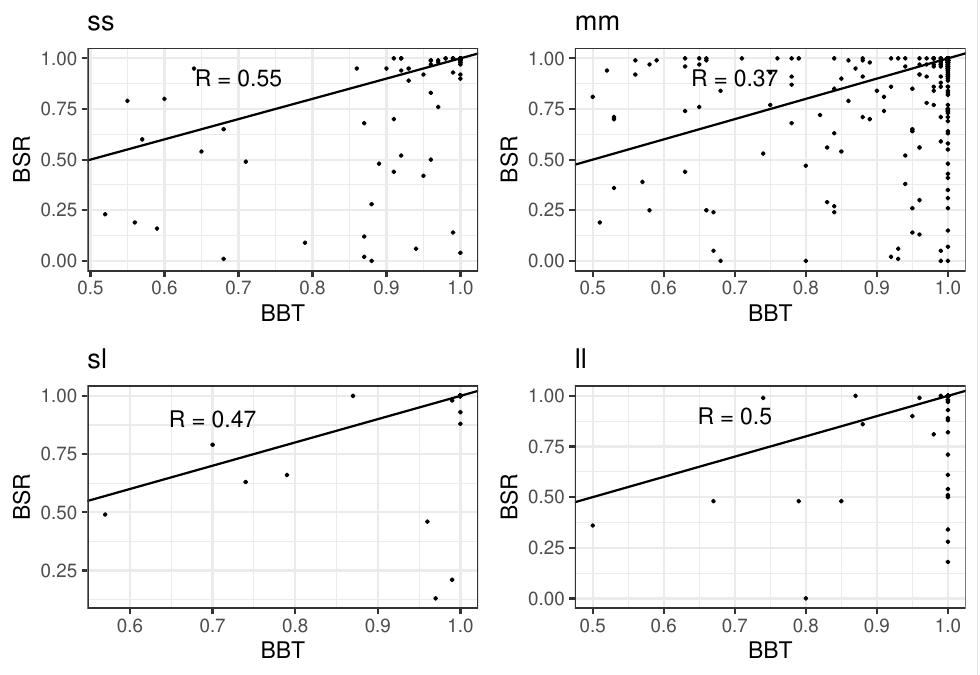} \caption{\label{fig:bsr}Comparison of the above0 from BSR (vertical axis) and {\em above.50} from BBT  (horizontal axis) for the ss, mm , sl and ll use cases.}\label{fig:bsr1}
\end{figure}

We believe that the only fair comparison is between the
{\em above.50} from BBT and the {\em above0} from BSR. Both measures
have the same semantics: {\em above0} counts the proportion of
evidence from BSR that one algorithm is better
than another and {\em above.50} under the weak interpretation also
counts the proportion of evidence that one algorithm is better than
another. Figure \ref{fig:bsr} plot the BSR {\em above0} measure against the
corresponding BBT {\em above.50} measure, where the dark line is the
\(y=x\) line. The figure also displays the Pearson correlation
coefficient of the two measures. As one can verify, there is a low
correlation between the results.

We do not know how to interpret the differences between the two
procedures, nor if the differences are important or not.

\begin{table}

\caption{\label{tab:xpairbays1}\label{tab:pairbays1}Results of the pairwise Bayesian Wilcoxon test. above.0 is the proportion of samples that indicates that the first algorithm is better than the second; in.rope is the proportion of evidence that falls wiith the 0.01 ROPE; and above.rope is the proportion of evidence that falls above the ROPE range.}
\centering
\begin{tabular}[t]{lrrr}
\toprule
\textbf{name} & \textbf{above0} & \textbf{in.rope} & \textbf{above.rope}\\
\midrule
lgbm > svm & 0.69 & 0.62 & 0.33\\
lgbm > xgb & 0.51 & 1.00 & 0.00\\
lgbm > dt & 1.00 & 0.00 & 1.00\\
lgbm > lda & 0.96 & 0.01 & 0.97\\
svm > xgb & 0.21 & 0.61 & 0.05\\
\addlinespace
svm > dt & 1.00 & 0.00 & 1.00\\
svm > lda & 0.98 & 0.03 & 0.94\\
xgb > dt & 1.00 & 0.00 & 1.00\\
xgb > lda & 0.97 & 0.02 & 0.96\\
dt > lda & 0.20 & 0.00 & 0.23\\
\bottomrule
\end{tabular}
\end{table}

\hypertarget{bbt-as-a-prediction}{%
\section{\texorpdfstring{BBT as a prediction \label{predictive}}{BBT as a prediction }}\label{bbt-as-a-prediction}}

Frequentist methods can be seen as decision procedures: given the data available, which algorithms can be said to be better than others. However, they do not make predictions about future data. The result of a frequentist test, such as a non-significant difference between two algorithms does not necessarily indicate that future data from new data sets will also not be significantly different. Given new data, the previously non-significant difference may become significant.

On the other hand, Bayesian methods can make predictions. For example, the output of the BBT model as shown in Table \ref{tab:main} can be considered as probabilities of one algorithm being better than the others for future data sets. There are two different predictions in the BBT model, as discussed in Section \ref{strong-weak}. In the strong interpretation, the {\em mean} measure predicts the proportion of wins for the better algorithm compared to the worse for future data sets. In the weak interpretation, the {\em above.50} measure predicts the proportion of wins for the better algorithm compared to the worse. In this section, we will test both of these predictions.

We conducted a series of experiments with the aim of determining the accuracy of predictions on future data sets. For each sample of training ss results, we sampled 10 additional data sets not used in the training sample, and evaluated the performance of each pair of algorithms on these test data sets. We counted the number of times one algorithm outperformed the other and referred to this count as ``win1'' for algorithm A1 and ``win2'' for algorithm A2. It is important to note that we did not apply local ROPE to the test data. The ratio \(win1/(win1+win2)\) is an empirical estimate of the probability that algorithm A1 is better than algorithm A2 on future test data sets.

For the strong interpretation, we examine the fit between the distribution of probability estimates \(P_s(1 \succ\,2)\) and the empirical probability \(win1/(win1+win2)\). Similar to the PPC summary table, we calculate the 50\%, 70\%, and 90\% highest density intervals (HDIs) of each distribution (for each pair of algorithms) and compare the proportion of empirical probabilities that fall within these intervals. In a well-calibrated distribution, at least 50\% of the empirical probabilities should fall within the 50\% HDI and so on. We also determine the proportion of empirical probabilities that fall outside of the 90\% HDI, both above its maximum value and below its minimum value. Finally, we calculate the mean error and the median absolute difference (MAD) between the mean prediction and the empirical probability.

For the weak interpretation, we do not have a distribution and therefore cannot use the same evaluation procedure as previously described. The {\em above.50} measure provides a probability statement, and we want to assess its accuracy. We will use the calibration plot procedure commonly used for classifiers. We divide the range of {\em above.50} values into three bins and compare the actual and expected number of cases where \(win1 > win2\). The actual number of cases is determined from the test data where {\em above.50} falls within the limits of the bin. The expected number of cases is the sum of {\em above.50} values in that bin. Typically, calibration plots divide probability estimates into 10 bins, but for this analysis, we have divided {\em above.50} into three bins: from 0.5 to 0.7, from 0.7 to 0.9, and from 0.9 to 1.0. This division represents low confidence (0.5 to 0.7), middle confidence (0.7 to 0.9), and high confidence (0.9 to 1.0) in the superiority of algorithm A over B.

Table \ref{tab:preds} shows the results of the strong interpretation evaluation. The predictions made by the BBT are not well-calibrated, as a much smaller proportion of empirical probabilities falls within the different HDIs than expected. In all cases, less than 50\% of the empirical probabilities fall within the 90\% HDI, which should contain 90\% of them. The values of {\em above90} and {\em below90} are somewhat similar, indicating that the BBT model is not systematically overestimating or underestimating the probability that one algorithm is better than another. The miss-calibration of the strong interpretation of the parameter is believed to be a problem of variance (incorrectly predicting the range of possible values) rather than a problem of bias (incorrectly predicting the most probable value). The mean prediction errors are low, ranging from 0.02 to 0.01, suggesting low bias. Therefore, it is concluded that the BBT is calibrated for its mean prediction, but too overconfident in the range of possible values, its credal interval.

The inclusion of the local ROPE was intended to serve both aesthetic and practical purposes. On one hand, it was included to address the issue of small differences that occur when computing average of cross-validations still counted as wins for one algorithm. On the other hand, local ROPE also aimed to reduce the model's overconfidence in its certainty regarding the estimates. The local ROPE reduces the number of wins for one algorithm over another, which should result in a decrease in the BBT's confidence in the probability that one algorithm is better than the other, thereby widening the credal interval. However, as Table \ref{tab:preds} indicates, the introduction of the local ROPE has limited impact in widening the credal interval, although it did have a small impact in reducing the error.

\begin{table}

\caption{\label{tab:predxz}\label{tab:preds}Prediction results - strong interpretation. Lrope and paired indicate whether local rope and the paried version of local rope was used. Within90 indicates the proportion of empirical win probability that falls within the 90\% HDI. Similarly for within70 and within50. Above90 is the proportion of empirical win probability that falls above the higher limit for the 90\% HDI; err is the mean error between the mean and the empirical win probability; and mad is the mean absolute error.}
\centering
\begin{tabular}[t]{ccccccccc}
\toprule
\textbf{lrope} & \textbf{paired} & \textbf{within.90} & \textbf{within.70} & \textbf{within.50} & \textbf{above90} & \textbf{below90} & \textbf{err} & \textbf{mad}\\
\midrule
\addlinespace[0.3em]
\multicolumn{9}{l}{\textbf{ss}}\\
\hspace{1em}F & F & 0.39 & 0.25 & 0.15 & 0.33 & 0.28 & 0.00 & 0.13\\
\hspace{1em}T & F & 0.44 & 0.25 & 0.14 & 0.33 & 0.23 & -0.01 & 0.12\\
\hspace{1em}T & T & 0.42 & 0.23 & 0.13 & 0.33 & 0.25 & -0.01 & 0.13\\
\addlinespace[0.3em]
\multicolumn{9}{l}{\textbf{mm}}\\
\hspace{1em}F & F & 0.33 & 0.22 & 0.16 & 0.37 & 0.31 & 0.00 & 0.08\\
\hspace{1em}T & F & 0.34 & 0.22 & 0.14 & 0.39 & 0.27 & -0.01 & 0.08\\
\hspace{1em}T & T & 0.32 & 0.22 & 0.16 & 0.41 & 0.27 & -0.01 & 0.08\\
\addlinespace[0.3em]
\multicolumn{9}{l}{\textbf{sl}}\\
\hspace{1em}F & F & 0.40 & 0.22 & 0.14 & 0.48 & 0.12 & -0.06 & 0.07\\
\hspace{1em}T & F & 0.38 & 0.26 & 0.14 & 0.50 & 0.12 & -0.07 & 0.07\\
\hspace{1em}T & T & 0.34 & 0.20 & 0.12 & 0.54 & 0.12 & -0.06 & 0.06\\
\bottomrule
\end{tabular}
\end{table}

\begin{table}

\caption{\label{tab:predxz2}\label{tab:predw}Prediction results - weak interpretation. Lrope and paired indicate whether local rope and the paried version of local rope was used. Pred50-70 is the expected number of examples whose empirical win  probability should fall within the [0.50,0.70] probability range; real50-70 is the observed number of empirical wins in that range. Simlarly for the other pairs of columns.}
\centering
\begin{tabular}[t]{cccccccc}
\toprule
\textbf{lrope} & \textbf{paired} & \textbf{pred50-70} & \textbf{real50-70} & \textbf{pred70-90} & \textbf{real70-90} & \textbf{pred90-100} & \textbf{real90-100}\\
\midrule
\addlinespace[0.3em]
\multicolumn{8}{l}{\textbf{ss}}\\
\hspace{1em}F & F & 4.3 & 3 & 13.8 & 11 & 75.3 & 64\\
\hspace{1em}T & F & 8.1 & 7 & 10.7 & 7 & 73.3 & 63\\
\hspace{1em}T & T & 6.1 & 4 & 11.1 & 8 & 75.2 & 65\\
\addlinespace[0.3em]
\multicolumn{8}{l}{\textbf{mm}}\\
\hspace{1em}F & F & 11.5 & 7 & 27.4 & 25 & 394.8 & 361\\
\hspace{1em}T & F & 9.9 & 8 & 25.2 & 20 & 398.4 & 370\\
\hspace{1em}T & T & 9.0 & 12 & 33.1 & 25 & 392.0 & 363\\
\addlinespace[0.3em]
\multicolumn{8}{l}{\textbf{sl}}\\
\hspace{1em}F & F & 1.9 & 2 & 0.0 & 0 & 46.6 & 45\\
\hspace{1em}T & F & 2.0 & 3 & 0.0 & 0 & 46.5 & 45\\
\hspace{1em}T & T & 2.0 & 3 & 0.0 & 0 & 46.5 & 45\\
\bottomrule
\end{tabular}
\end{table}

Table \ref{tab:predw} presents the results of the weak interpretation calibration. The results indicate better calibration compared to the strong interpretation. The predictions for the range of 0.5 to 0.7 are few and align closely with the empirical results for all three use cases. Similarly, the predictions for the range of 0.7 to 0.9 are accurate. However, for the high confidence range of 0.9 to 1.0, the predictions appear to be slightly overconfident, slightly higher than the empirical value. The BBT model seems to have an over-confidence in its predictions, and the introduction of the local ROPE and paired local ROPE had no effect on reducing this over-confidence.

\hypertarget{discussion}{%
\section{\texorpdfstring{Discussion \label{sec:disc}}{Discussion }}\label{discussion}}

BBT, as a comparison procedure, has several advantages over some traditional frequentist approaches. Despite making a binary decision of accepting or rejecting the difference between two algorithms in an aggregated ranking, BBT seems to be more powerful than Demsar's procedure and more robust than pairwise Wilcoxon-based approaches. This results in BBT making more decisions of ``significant'' differences compared to these two frequentist tests. As demonstrated by \citet{benavoli2017time}, the BSR procedure also makes more decisions than Demsar's when comparing only pairs of classifiers.

Comparing BBT with BSR is challenging due to the lack of agreement between the results generated by the two procedures. However, BBT addresses several limitations of BSR, including the ability to handle multiple comparisons, the ROPE in BSR being valid only for accuracy and not other comparable metrics, and the inability to compare incomparable metrics.

The Bradley-Terry formalis seems simple but sufficiently complex to model a comparison of machine learning algorithms on multiple data. The predictive posterior check show that the Bayesian
model is indeed a good model of the data that was given, and it showed
that a more complex model such as Davidson's is not needed and it
worsens the fitness between the model and the data. The hyper prior
tests (Section \ref{test-hyper}) show that the model is not sensitive
to different reasonable hyper priors, and the test on different
policies to deal with ties (Section \ref{test-ties}), besides using
the Davidson model, are basically equivalent. All this should point to
the conclusion that the model is stable to different decisions, and it
is generally a good fit to the data.

As we already mentioned, the mode is simple, the MCMC converges well
with few samples, and our implementation (thanks to Stan's MCMC) runs
in less than a second in a modern laptop.

Regarding the ``predictive'' part of the model it is yet unclear whether
the apparent overconfidence of the model, specially under the strong
interpretation is a problem. Since no frequentist test can make
predictions, and the BSR did not test its predictive fitness, we do
not have an alternative to compare against.

\hypertarget{code-and-data-availability}{%
\subsection{\texorpdfstring{Code and data availability \label{sec:code}}{Code and data availability }}\label{code-and-data-availability}}

An R package (bbtcomp) that implements the BBT model is available at
\url{https://github.com/jwainer/bbtcomp}. To install it use
\texttt{remotes::install\_github("jwainer/bbtcomp")}. The R package uses the
\texttt{cmdstanr} package to interface with Stan, which implements the MCMC
sampler. At the time of the writing, \texttt{cmdstanr} is not available in
CRAN, and should be installed following the instructions in
\url{https://mc-stan.org/cmdstanr/}.

A Python program that implements all functionalities of the R package
implementation the BBT model with the exception of the graphic
generating functions, is available in the github directory above, in
the folder \texttt{python}. The program also uses the \texttt{cmdstanpy} interface to Stan.

The programs that generated the examples and experiments in this paper, as
well as the Rmarkdown version of this paper is available in the github
directory above, in the folder \texttt{paper}.

\hypertarget{how-to-use-the-bbt-model---weak-interpretation-and-0.95-probabilities}{%
\subsection{How to use the BBT model - weak interpretation and 0.95 probabilities}\label{how-to-use-the-bbt-model---weak-interpretation-and-0.95-probabilities}}

We believe that there are two main approaches to utilizing the BBT model in research. For researchers or audiences who are more familiar with the frequentist approach, we recommend using the summary values related to the weak interpretation of the parameters (the {\em above.50} and {\em in.rope} values) and a 0.95 probability threshold.

The guidelines for this approach are as follows:

\begin{itemize}
\item
  If the {\em in.rope} value is 0.95 or above, the researchers can claim that the two algorithms are equivalent (according to the definition of ROPE).
\item
  If {\em above.50} is 0.95 or above, the researchers can claim that one algorithm is better than the other. If both {\em in.rope} and {\em above.50} are above 0.95, the first rule applies, and researchers should claim that the algorithms are equivalent.
\item
  For all other cases, researchers should not make any claim.
\end{itemize}

This approach utilizes familiar threshold numbers such as 0.95 or 95\%, and, as seen in section 5, it should provide results that are stronger than Demsar's procedure and the pairwise Wilcoxon procedure. It will identify more pairwise comparisons as ``significant'' while still not missing any comparisons deemed significant by Demsar's procedure or the Wilcoxon procedure (section \ref{sec:comp-wil}).

The BBT model also allows one to make claims of equivalence (when
{\em in.rope} \(\ge 0.95\)) that are not possible in the frequentist
case.

Regarding the explanatory aspect of the procedure, for the pairs of
algorithms with ``significant'' differences, the researcher can claim
that for the \emph{data given}, 95\% or more of the estimates of the
relative strengths of the algorithms indicate that algorithm A is
better than algorithm B.

Finally, the BBT model is reasonably well-calibrated although slightly
overconfident regarding high values of {\em above.50}, so for the
pairs with ``significant'' differences the researcher can make the claim
that, \emph{likely} with 90\% probability or better, the best algorithm
should perform better than the worse for \emph{future data sets}.

\hypertarget{how-to-use-the-bbt-model---strong-interpretation-and-0.70-probabilities}{%
\subsection{How to use the BBT model - strong interpretation and 0.70 probabilities}\label{how-to-use-the-bbt-model---strong-interpretation-and-0.70-probabilities}}

If researchers and their audiences are more comfortable with Bayesian results, we recommend following the strong interpretation (and the {\em mean} summary of the probabilities) and choosing a threshold of 0.70. The procedure is as follows:

\begin{itemize}
\item
  If {\em mean} is below 0.55, one can claim that both algorithms are equivalent.
\item
  If {\em mean} is above 0.70, one can claim that one algorithm is ``significantly'' better than the other.
\end{itemize}

In terms of explanation, the researcher can state that, based on the \emph{given data}, the mean estimate of the probability that algorithm A is better than B is at least 70\%, with the 89\% (or 95\%, if preferred) of the estimates falling within the {\em low} to {\em high} credal interval.

For \emph{future data}, the researcher can make the claim that the \emph{most likely} value for the probability that A is better than B is 70\% or better. The low predictive bias allows for such a claim, but the high variance does not allow for a credal interval to be defined for these estimates.

The strong interpretation with a 0.70 threshold may be comparable to Demsar's results for small-scale use cases, but it becomes a more stringent criterion for larger-scale cases. Some pairs that are determined to be significantly different in Demsar's procedure may not be considered as such using the 0.70 threshold. For the repeated experiments:

\begin{itemize}
\item
  for \textbf{ss} use cases: the 0.70 threshold detects 18 new pairs of
  comparisons (out of 100) that Demsar's procedure does not find
  significantly different, and detects 9 pairs that the pairwise
  Wilcoxon does not find significantly different, but does not detect
  5 that the pairwise Wilcoxon does.
\item
  for \textbf{mm} use cases: the 0.70 threshold finds 10 new pairs (out of
  100), but misses 7 in comparison to Demsar, and finds new 27 but
  misses 73 in comparison to pairwise Wilcoxon.
\item
  for \textbf{sl} use cases: the 0.70 threshold finds no new pairs, but
  misses 11 (out of 50) in comparison to Demsar's, and also misses 17
  in comparison to the pairwise Wilcoxon.
\item
  for the $\ell\ell$-results: the 0.70 threshold misses 16 (out of 120)
  pairs in comparison to Demsar, and misses 38 in comparison to
  Wilcoxon.
\end{itemize}

Therefore, using the 0.70 strong interpretation threshold will result in fewer pairs being flagged as ``different'' as the size of the comparisons increases. However, unlike any frequentist-based ``significantly different'' pair of algorithms, one can still claim that for those flagged pairs in the BBT procedure, the best algorithm in each pair will likely perform better than the other one in 70\% of future data sets.

Finally, as discussed in section \ref{sec:xextra}, the strong interpretation is somewhat independent of the number of algorithms being compared, and thus can be used in these situations.

\hypertarget{conclusion}{%
\section{\texorpdfstring{Conclusion \label{sec:conc}}{Conclusion }}\label{conclusion}}

The BBT model is a comparison procedure based on the Bradley-Terry model that assigns a merit value to each of the K competing algorithms. The merit values determine the probability of an algorithm performing better than another on a data set. The BBT model is a Bayesian implementation of the Bradley-Terry model and offers several advantages over traditional frequentist approaches:

\begin{itemize}
\item
  It allows for a more nuanced description of the relationships between each pair of algorithms in the aggregated ranking, beyond simply determining whether the difference is significant or not.
\item
  It enables the definition of a threshold below which two algorithms are considered equivalent for practical purposes and provides a means to make claims about whether two algorithms are equivalent or not.
\item
  It provides an understanding of the uncertainties associated with the claims.
\end{itemize}

In addition to the advantages of Bayesian estimation, the BBT model also:

\begin{itemize}
\item
  Works with any metric of interest, regardless of whether it is comparable or not.
\item
  The main parameters are probabilities, making it easier to understand the definition of ROPEs, uncertainties, and so on.
\item
  Accommodates missing data for algorithms that did not run on certain data sets.
\item
  Does not need different statistical procedures for all pairwise comparisons or comparisons against a control.
\end{itemize}

Finally, we also introduced the idea of local ROPE, which is a method for determining when one algorithm can be considered truly better than another for a specific data set, based on their average performance across different folds. We believe that local ROPE can also be a useful addition to frequentist tests, particularly rank-based tests.

This paper did not dedicate extensive space to the defense of Bayesian testing methods. For a more in-depth examination of the advantages of Bayesian testing, we refer the reader to the work of \citet{benavoli2017time}, which not only presents the BSR model but also makes compelling arguments for the machine learning community to transition away from frequentist tests and towards Bayesian tests. The author of this paper agrees with these arguments and recommendations.

\appendix

\hypertarget{alternative-hyper-priors}{%
\section{\texorpdfstring{Alternative hyper-priors \label{test-hyper}}{Alternative hyper-priors }}\label{alternative-hyper-priors}}

The BBT Model uses a lognormal distribution with a 0.5 scale parameter
as a hyper-prior for the \(\sigma\) parameter, as proposed in \citet{btstan}. But
that author also suggested using different scales for the lognormal
distribution as well as a Cauchy distribution (with scale = 1.0). On
the other hand, \citet{issa2021bayesian} propose a Normal distribution with
standard deviation 3.0 as that hyper-prior. We tested these
alternatives and they make no important difference. We performed the
usual repeated experiments, measuring WAIC and the calibration of the
PPC as in Tables \ref{tab:hyper1} and \ref{tab:hyper1b}

\begin{table}

\caption{\label{tab:hyperxz1}\label{tab:hyper1}The PPC of ss and mm use cases for the different hyper-priors. Hyper indicates the hyper-prior distribution, scale its scale parameter (mean or similar parameter). The other columns are the usual columns for the PPC table.}
\centering
\begin{tabular}[t]{lrrrrrrrrrrr}
\toprule
\multicolumn{2}{c}{ } & \multicolumn{5}{c}{ss} & \multicolumn{5}{c}{mm} \\
\cmidrule(l{3pt}r{3pt}){3-7} \cmidrule(l{3pt}r{3pt}){8-12}
\textbf{\textbf{hyper}} & \textbf{\textbf{scale}} & \textbf{\textbf{waic}} & \textbf{\textbf{h50}} & \textbf{\textbf{h90}} & \textbf{\textbf{h95}} & \textbf{\textbf{h100}} & \textbf{\textbf{waic}} & \textbf{\textbf{h50}} & \textbf{\textbf{h90}} & \textbf{\textbf{h95}} & \textbf{\textbf{h100}}\\
\midrule
cauchy & 0.50 & 40.85225 & 0.83 & 1 & 1 & 1 & 219.7626 & 0.728 & 0.998 & 1 & 1\\
cauchy & 1.00 & 40.76751 & 0.83 & 1 & 1 & 1 & 219.7257 & 0.715 & 0.996 & 1 & 1\\
lognormal & 0.25 & 40.58881 & 0.87 & 1 & 1 & 1 & 219.6236 & 0.708 & 0.996 & 1 & 1\\
lognormal & 0.50 & 40.56506 & 0.90 & 1 & 1 & 1 & 219.6175 & 0.699 & 0.996 & 1 & 1\\
lognormal & 1.00 & 40.73847 & 0.85 & 1 & 1 & 1 & 219.6493 & 0.728 & 0.996 & 1 & 1\\
\addlinespace
normal & 2.00 & 40.71471 & 0.84 & 1 & 1 & 1 & 219.5983 & 0.722 & 0.994 & 1 & 1\\
normal & 5.00 & 40.69713 & 0.86 & 1 & 1 & 1 & 219.6517 & 0.705 & 0.996 & 1 & 1\\
\bottomrule
\end{tabular}
\end{table}

\begin{table}

\caption{\label{tab:hyperxz2}\label{tab:hyper1b}The PPC of sl and ll use cases the different hyper-priors. Hyper indicates the hyper-prior distribution, scale its scale parameter (mean or similar parameter). The other columns are the usual columns for the PPC table.}
\centering
\begin{tabular}[t]{lrrrrrrrrrrr}
\toprule
\multicolumn{2}{c}{ } & \multicolumn{5}{c}{sl} & \multicolumn{5}{c}{ll} \\
\cmidrule(l{3pt}r{3pt}){3-7} \cmidrule(l{3pt}r{3pt}){8-12}
\textbf{\textbf{hyper}} & \textbf{\textbf{scale}} & \textbf{\textbf{waic}} & \textbf{\textbf{h50}} & \textbf{\textbf{h90}} & \textbf{\textbf{h95}} & \textbf{\textbf{h100}} & \textbf{\textbf{waic}} & \textbf{\textbf{h50}} & \textbf{\textbf{h90}} & \textbf{\textbf{h95}} & \textbf{\textbf{h100}}\\
\midrule
cauchy & 0.50 & 62.51531 & 0.64 & 1.00 & 1 & 1 & 767.2083 & 0.48 & 0.92 & 0.96 & 1\\
cauchy & 1.00 & 62.33314 & 0.66 & 1.00 & 1 & 1 & 766.7715 & 0.47 & 0.92 & 0.96 & 1\\
lognormal & 0.25 & 62.36310 & 0.68 & 1.00 & 1 & 1 & 767.1563 & 0.46 & 0.92 & 0.97 & 1\\
lognormal & 0.50 & 62.50422 & 0.62 & 1.00 & 1 & 1 & 767.2467 & 0.46 & 0.92 & 0.96 & 1\\
lognormal & 1.00 & 62.59410 & 0.62 & 0.98 & 1 & 1 & 767.6128 & 0.48 & 0.93 & 0.96 & 1\\
\addlinespace
normal & 2.00 & 62.45112 & 0.64 & 1.00 & 1 & 1 & 767.7501 & 0.47 & 0.92 & 0.97 & 1\\
normal & 5.00 & 62.54199 & 0.70 & 1.00 & 1 & 1 & 768.0781 & 0.48 & 0.92 & 0.96 & 1\\
\bottomrule
\end{tabular}
\end{table}

\newpage

\vskip 0.2in
\bibliography{ref.bib}

\end{document}

%% file: twin1.tex
\begin{tabular}{llrrr}
  \toprule
{\bfseries alg1} & {\bfseries alg2} & {\bfseries win1} & {\bfseries win2} & {\bfseries ties} \\ 
  \midrule
dt & lda & 6 & 13 & 1 \\ 
  dt & lgbm & 0 & 17 & 3 \\ 
  dt & xgb & 0 & 17 & 3 \\ 
  dt & svm & 5 & 14 & 1 \\ 
  lda & lgbm & 6 & 13 & 1 \\ 
  lda & xgb & 5 & 14 & 1 \\ 
  lda & svm & 5 & 15 & 0 \\ 
  lgbm & xgb & 9 & 8 & 3 \\ 
  lgbm & svm & 10 & 9 & 1 \\ 
  xgb & svm & 11 & 8 & 1 \\ 
   \bottomrule
\end{tabular}

%% file: twin2.tex
\begin{tabular}{llrr}
  \toprule
{\bfseries alg1} & {\bfseries alg2} & {\bfseries win1} & {\bfseries win2} \\ 
  \midrule
dt & lda & 7 & 14 \\ 
  dt & lgbm & 2 & 19 \\ 
  dt & xgb & 2 & 19 \\ 
  dt & svm & 6 & 15 \\ 
  lda & lgbm & 7 & 14 \\ 
  lda & xgb & 6 & 15 \\ 
  lda & svm & 5 & 15 \\ 
  lgbm & xgb & 11 & 10 \\ 
  lgbm & svm & 11 & 10 \\ 
  xgb & svm & 12 & 9 \\ 
   \bottomrule
\end{tabular}

%% file: miss1.tex
\begin{tabular}{llrr}
  \toprule
{\bfseries alg1} & {\bfseries alg2} & {\bfseries win1} & {\bfseries win2} \\ 
  \midrule
dt & lda & 7 & 14 \\ 
  dt & lgbm & 2 & 19 \\ 
  dt & xgb & 2 & 17 \\ 
  dt & svm & 6 & 15 \\ 
  lda & lgbm & 7 & 14 \\ 
  lda & xgb & 6 & 13 \\ 
  lda & svm & 5 & 15 \\ 
  lgbm & xgb & 10 & 9 \\ 
  lgbm & svm & 11 & 10 \\ 
  xgb & svm & 10 & 9 \\ 
   \bottomrule
\end{tabular}

%% file: miss2.tex
\begin{tabular}{lrrrr}
  \toprule
{\bfseries pair} & {\bfseries mean} & {\bfseries delta} & {\bfseries above.50} & {\bfseries in.rope} \\ 
  \midrule
lgbm $>$ xgb & 0.51 & 0.24 & 0.55 & 0.51 \\ 
  lgbm $>$ svm & 0.54 & 0.22 & 0.74 & 0.45 \\ 
  lgbm $>$ lda & 0.70 & 0.19 & 1.00 & 0.01 \\ 
  lgbm $>$ dt & 0.82 & 0.15 & 1.00 & 0.00 \\ 
  xgb $>$ svm & 0.53 & 0.23 & 0.69 & 0.46 \\ 
  xgb $>$ lda & 0.69 & 0.20 & 1.00 & 0.02 \\ 
  xgb $>$ dt & 0.81 & 0.15 & 1.00 & 0.00 \\ 
  svm $>$ lda & 0.66 & 0.20 & 0.99 & 0.04 \\ 
  svm $>$ dt & 0.79 & 0.17 & 1.00 & 0.00 \\ 
  lda $>$ dt & 0.66 & 0.21 & 0.99 & 0.05 \\ 
   \bottomrule
\end{tabular}

%% file: tdemsar1.tex
\begin{tabular}{ll}
  \toprule
{\bfseries name} & {\bfseries dif} \\ 
  \midrule
xgb $>$ lgbm &   \\ 
  xgb $>$ svm &   \\ 
  xgb $>$ lda &   \\ 
  xgb $>$ dt & yes \\ 
  lgbm $>$ svm &   \\ 
  lgbm $>$ lda &   \\ 
  lgbm $>$ dt & yes \\ 
  svm $>$ lda &   \\ 
  svm $>$ dt & yes \\ 
  lda $>$ dt &   \\ 
   \bottomrule
\end{tabular}

%% file: twilc2.tex
\begin{tabular}{lr}
  \toprule
{\bfseries algorithm} & {\bfseries median} \\ 
  \midrule
lgbm & 0.93 \\ 
  svm & 0.93 \\ 
  xgb & 0.93 \\ 
  dt & 0.87 \\ 
  lda & 0.85 \\ 
   \bottomrule
\end{tabular}

%% file: twilc1.tex
\begin{tabular}{lr}
  \toprule
{\bfseries name} & {\bfseries p.value} \\ 
  \midrule
lgbm $>$ svm & 0.98 \\ 
  lgbm $>$ xgb & 0.98 \\ 
  lgbm $>$ dt & 0.00 \\ 
  lgbm $>$ lda & 0.47 \\ 
  svm $>$ xgb & 0.98 \\ 
  svm $>$ dt & 0.11 \\ 
  svm $>$ lda & 0.46 \\ 
  xgb $>$ dt & 0.00 \\ 
  xgb $>$ lda & 0.46 \\ 
  dt $>$ lda & 0.98 \\ 
   \bottomrule
\end{tabular}